\documentclass{CVM}
\usepackage{lmodern}
\usepackage{anyfontsize}

\usepackage{amsmath}
\usepackage{graphicx}
\usepackage{textcomp}
\usepackage{subcaption}
\usepackage{tikz}
\usepackage{tabularx}
\usepackage{booktabs}
\usepackage{multirow}
\usepackage{pifont} 
\usepackage[colorlinks=true, 
            linkcolor=blue, 
            citecolor=violet, 
            urlcolor=blue]{hyperref}
\usepackage{colortbl}
\usepackage{graphicx}
\usepackage{booktabs}
\usepackage{adjustbox}
\usepackage{array}
\usepackage{makecell}

\usepackage{caption} 

\CVMsetup{
type      = {Review Article},
doi       = {CVM.XXXX},  % Replace with actual DOI if available
title     = {A Survey of 3D Reconstruction with Event Cameras},
author    = {Chuanzhi Xu$^1$, Haoxian Zhou$^1$, Langyi Chen$^1$, Haodong Chen$^1$, Zeke Zexi Hu$^1$, Zhicheng Lu$^2$, Ying Zhou$^1$, Vera Chung$^1$, Qiang Qu$^1$\cor{}, and Weidong Cai$^1$},
runauthor = {C. Xu et al.},
abstract  = {
Event cameras are rapidly emerging as powerful vision sensors for 3D reconstruction, uniquely capable of asynchronously capturing per-pixel brightness changes. Compared to traditional frame-based cameras, event cameras produce sparse yet temporally dense data streams, enabling robust and accurate 3D reconstruction even under challenging conditions such as high-speed motion, low illumination, and extreme dynamic range scenarios. These capabilities offer substantial promise for transformative applications across various fields, including autonomous driving, robotics, aerial navigation, and immersive virtual reality. In this survey, we present the first comprehensive review exclusively dedicated to event-based 3D reconstruction. Existing approaches are systematically categorised based on input modality into stereo, monocular, and multimodal systems, and further classified according to reconstruction methodologies, including geometry-based techniques, deep learning approaches, and neural rendering techniques such as Neural Radiance Fields (NeRF) and 3D Gaussian Splatting (3DGS). Within each category, methods are chronologically organised to highlight the evolution of key concepts and advancements. Furthermore, we provide a detailed summary of publicly available datasets specifically suited to event-based reconstruction tasks. Finally, we discuss significant open challenges in dataset availability, standardised evaluation, effective representation, and dynamic scene reconstruction, outlining insightful directions for future research. This survey aims to serve as an essential reference and provides a clear and motivating roadmap toward advancing the state of the art in event-driven 3D reconstruction. \href{https://github.com/LouckXu/Event3DSurvey}{Project Page}
},
keywords  = {3D Reconstruction, Event Camera, Neuromorphic Vision, Event-based Vision, Neural Radiance Fields, 3D Gaussian Splatting},
copyright = {The Author(s)},
}

\begin{document}
\maketitle

\enlargethispage{-4pt}
\begin{figure}[b] \vskip -2mm
\small\renewcommand\arraystretch{0.85}
\begin{tabular}{p{80.5mm}} \toprule\\ \end{tabular}
\vskip -2.5mm \noindent \setlength{\tabcolsep}{1pt}
\begin{tabular}{p{3.5mm}p{80mm}}
$1\quad $ & School of Computer Science, The University of Sydney, NSW 2006, Australia.

E-mail: 
{Chuanzhi Xu, \href{mailto:chuanzhi.xu@sydney.edu.au}{chuanzhi.xu@sydney.edu.au};

Haoxian Zhou, \href{mailto:hzho0442@uni.sydney.edu.au}{hzho0442@uni.sydney.edu.au};

Langyi Chen, \href{mailto:lche5181@uni.sydney.edu.au}{lche5181@uni.sydney.edu.au};

Haodong Chen, \href{mailto:haodong.chen@sydney.edu.au}{haodong.chen@sydney.edu.au};

Zeke Zexi Hu, \href{mailto:zexi.hu@sydney.edu.au}{zexi.hu@sydney.edu.au};

Ying Zhou, \href{mailto:ying.zhou@sydney.edu.au}{ying.zhou@sydney.edu.au};

Vera Chung, \href{mailto:vera.chung@sydney.edu.au}{vera.chung@sydney.edu.au};

Qiang Qu\cor{}, \href{mailto:vincent.qu@sydney.edu.au}{vincent.qu@sydney.edu.au};

Weidong Cai, \href{mailto:tom.cai@sydney.edu.au}{tom.cai@sydney.edu.au}
} \\ [1mm]

$2\quad $ & Rural Health Research Institute, Charles Sturt University, Orange, NSW, Australia.

E-mail:
{Zhicheng Lu, \href{mailto:zlu@csu.edu.au}{zlu@csu.edu.au}
} \\

&\hspace{-5mm} Manuscript submitted: 2025-06; accepted: 2025-12\vspace{-2mm}
\end{tabular} \vspace {-3mm}
\end{figure}

\makeatletter
\def\ps@accepted{%
  \def\@oddhead{%
    \hspace*{0.9\linewidth}%
    \large\textcolor{gray}{This survey has been accepted to Computational Visual Media}%
    \hfil
  }%
  \def\@evenhead{\@oddhead}%
  \def\@oddfoot{}%
  \def\@evenfoot{}%
}
\makeatother
\thispagestyle{accepted}

\section{Introduction}
\begin{figure*}[th!]
    \centering
    \includegraphics[width=\linewidth]{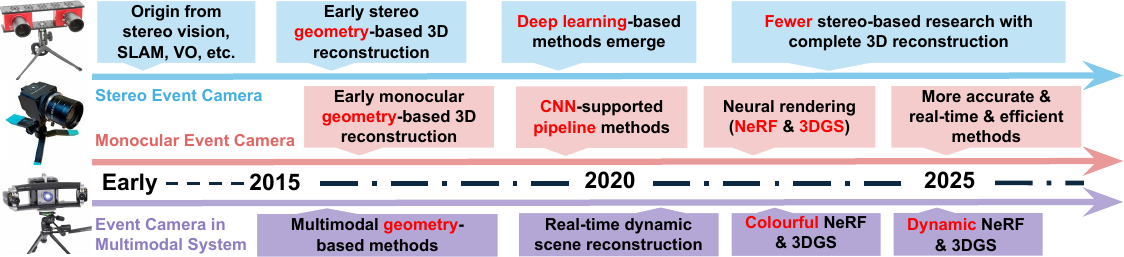}
    \caption{\textbf{Roadmap of 3D reconstruction with event cameras.} It shows the development from event-based geometry to neural 3D rendering. With advances in technology, event-camera-based 3D reconstruction methods are achieving progressively higher accuracy and realism, enabling more complete and faithful 3D scene rendering.}
    \label{fig:roadmap}
\end{figure*}

\begin{figure*}[t]
  \centering
  \includegraphics[width=0.7\textwidth]{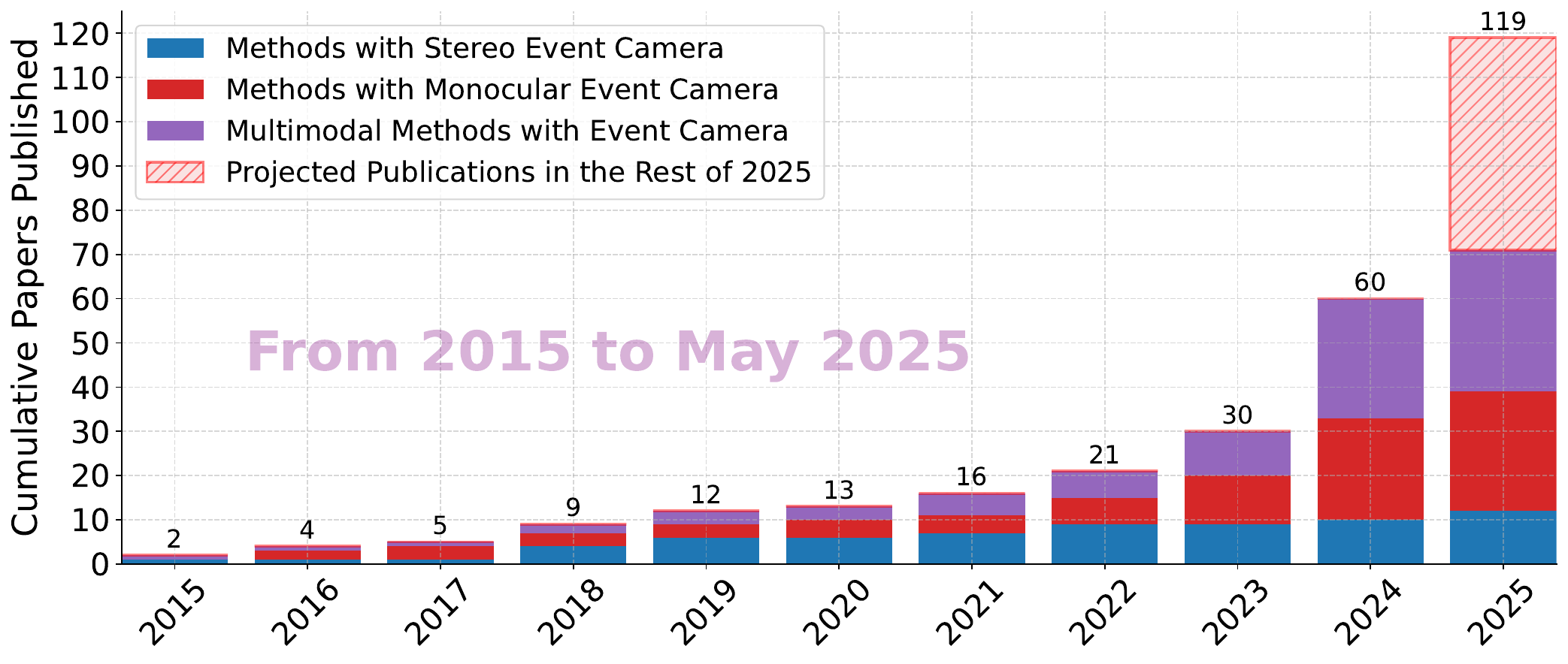}
  \caption{Publication trends by category (May 2025)}
  \label{fig:cumulative}
\end{figure*}

\textbf{3D reconstruction} is a fundamental technique that transforms 2D observations of real-world scenes, objects, or simulated environments into accurate 3D models \cite{3dReconstruction}. Typical outputs from these reconstruction processes include depth maps \cite{saxena2008make3d}, point clouds \cite{PointCloud}, meshes \cite{Mesh}, and voxel representations \cite{Voxel}. Numerous sensing technologies are used to capture the essential data required for 3D reconstruction, including conventional RGB cameras \cite{RGB3drecon}, RGB-D cameras \cite{li2022high}, structured light sensors \cite{StructureLightSurvey}, and LiDAR systems \cite{LiDarSurvey, 3DSurvey}. However, each of these traditional sensors has inherent limitations: conventional RGB cameras struggle under extreme lighting conditions and rapid motion scenarios, whereas active sensors such as LiDAR and structured-light systems are typically bulky and require substantial physical space \cite{3DSurvey}. Consequently, event cameras have emerged as an attractive alternative or complementary sensing modality, addressing many of these drawbacks.

\textbf{Event cameras}, also referred to as neuromorphic cameras, silicon retinas, or dynamic vision sensors, are bio-inspired sensors designed to asynchronously capture brightness changes rather than producing images at a fixed frame rate \cite{lichtsteiner2008128}. Unlike conventional RGB cameras, each pixel in an event camera independently detects and reports intensity changes, effectively acting as an individual sensor element \cite{posch2014retinomorphic}. When the brightness at a pixel exceeds a predefined threshold, the event camera generates event data consisting of the pixel’s coordinates, timestamp, and the polarity of the brightness change. Due to their unique sensing principle, event cameras exhibit outstanding advantages, such as high temporal resolution, low latency, and robustness to motion blur and challenging lighting conditions \cite{lichtsteiner2008128}. These properties have driven extensive research on event cameras across various vision tasks, including object detection \cite{Detection}, recognition \cite{Recognition}, segmentation \cite{Segmentation}, and tracking \cite{Tracking}; video enhancement tasks such as super-resolution \cite{SuperResolution}, frame interpolation \cite{Interpolation}, and deblurring \cite{Deblurring}; as well as complex spatiotemporal modeling tasks including simultaneous localisation and mapping (SLAM) \cite{SLAM}, visual odometry (VO) \cite{VO}, and notably, 3D reconstruction. Event-based vision technologies have demonstrated considerable potential in diverse applications, ranging from autonomous driving \cite{AutoDriving}, robotics \cite{Robot}, and unmanned aerial vehicles (UAVs) \cite{UAV} to industrial monitoring \cite{Industry}, etc.

Leveraging their distinct sensing capabilities, event cameras have gained significant attention in recent years for performing 3D reconstruction. It is a domain traditionally dominated by standard cameras since the 1990s \cite{[R1]}. Event-based 3D reconstruction approaches now span from occupancy mapping \cite{[Towards]} and real-time dynamic scene reconstruction \cite{[M3]}, to achieving high-fidelity rendering of complex scenes \cite{[G1]}. Figure \ref{fig:roadmap} illustrates the roadmap and evolution of event-based 3D reconstruction techniques. Initial efforts utilising event cameras for 3D reconstruction began emerging prominently in the 2010s \cite{[2013], [S100]}. Existing reconstruction strategies employ stereo and monocular event camera setups \cite{[S9], [M1]}, alongside multimodal approaches integrating event data with other sensory modalities \cite{[DMT1]}. Recent advances further integrate sophisticated neural rendering methods, such as Neural Radiance Fields (NeRF) and 3D Gaussian Splatting (3DGS), to enable unprecedented levels of realism \cite{[N3], [G1]}.

\begin{figure*}[ht!]
  \centering
  \includegraphics[width=0.7\textwidth]{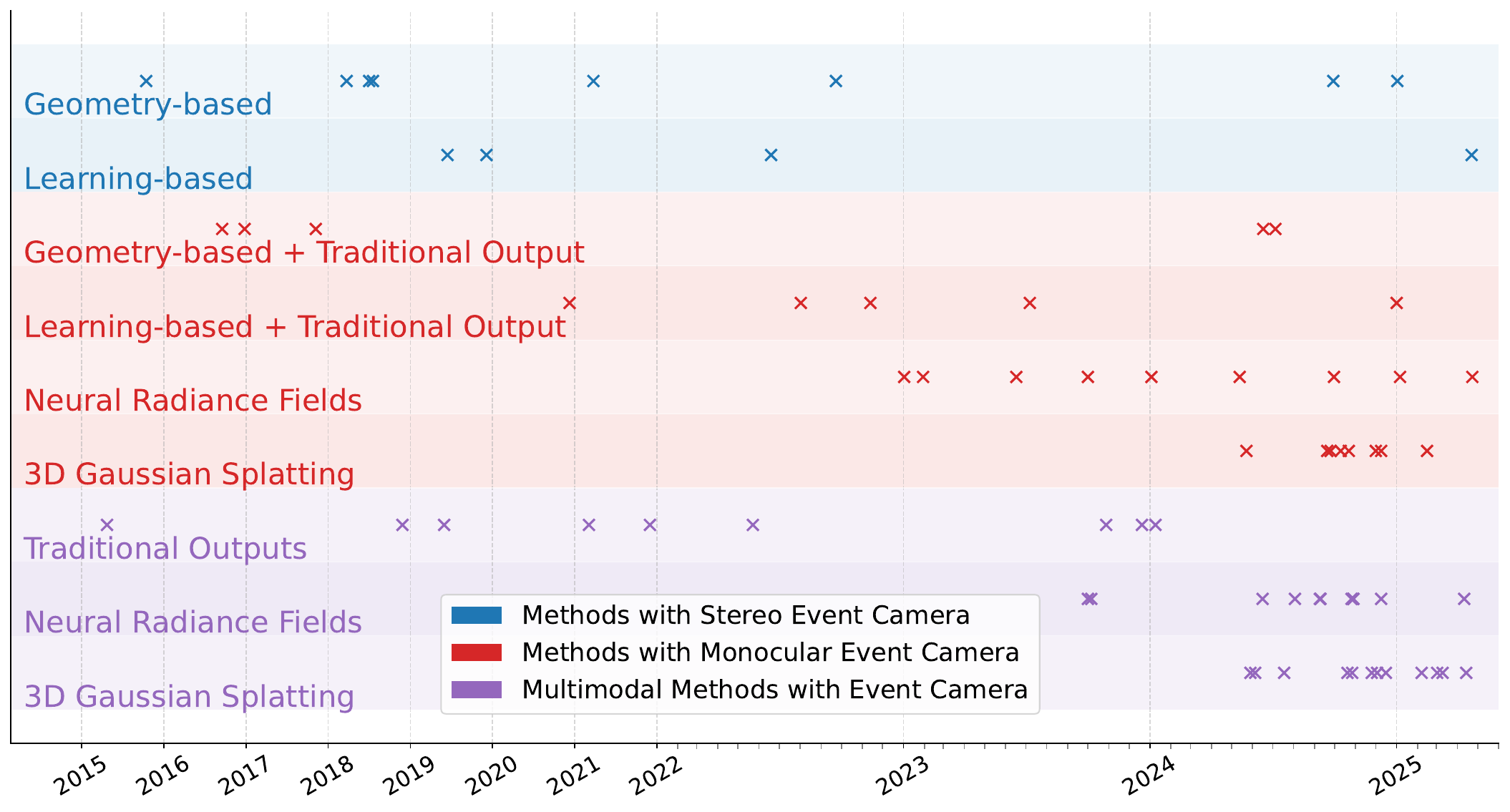}
  \caption{Publication timeline in categories (from 2015 to May 2025). The research focus has shifted from early geometric and traditional learning methods to NeRF and 3D Gaussian Splatting, particularly in monocular and multimodal settings, where these emerging techniques have become dominant since 2023.}
  \label{fig:timeline}
\end{figure*}

Notably, a complete pipeline from raw event data acquisition to final 3D model reconstruction should be able to output a representable 3D structure, either in the form of a directly modellable 3D representation or at least a depth map that can be readily converted into a 3D point cloud (as in some depth estimation methods \cite{[S4], [S0]}). However, numerous earlier works that lack a final 3D output, such as those in SLAM \cite{SLAM}, VO \cite{VO}, and 3D perception \cite{[Non3d]}, can be considered precursor steps toward 3D reconstruction, but should not be regarded as complete event-to-3D model reconstruction pipelines. In Figure \ref{fig:cumulative}, publications dedicated to event-based 3D reconstruction have significantly increased since 2015, experiencing rapid growth particularly in the last three years, underscoring the burgeoning interest and research value in this area. Since the choice of sensing configuration heavily influences the subsequent reconstruction pipeline, existing methods are typically categorised according to the camera setup: stereo, monocular, and multimodal approaches.

Despite the steadily growing number of research publications, there currently exists no dedicated survey for event-driven 3D reconstruction. Three comprehensive but broader surveys have briefly touched upon aspects of 3D reconstruction using event cameras. Gallego et al. (2020) provided a wide-ranging overview of event cameras \cite{[R1]}, which included a brief section on 3D reconstruction. However, rapid advances have rendered parts of that review outdated. More recently, Chakravarthi et al. (2024) categorised event camera research tasks \cite{[R2]}, offering only limited insights into 3D reconstruction. Similarly, Zheng et al. (2024) discussed deep learning methods for event cameras \cite{[R3]}, with just a brief subsection dedicated to 3D reconstruction and limited historical context. Therefore, an explicit, comprehensive, and updated survey is essential to systematically explore and evaluate the advancements in event-driven 3D reconstruction.

\begin{figure*}[th!]
    \centering
    \includegraphics[width=\linewidth]{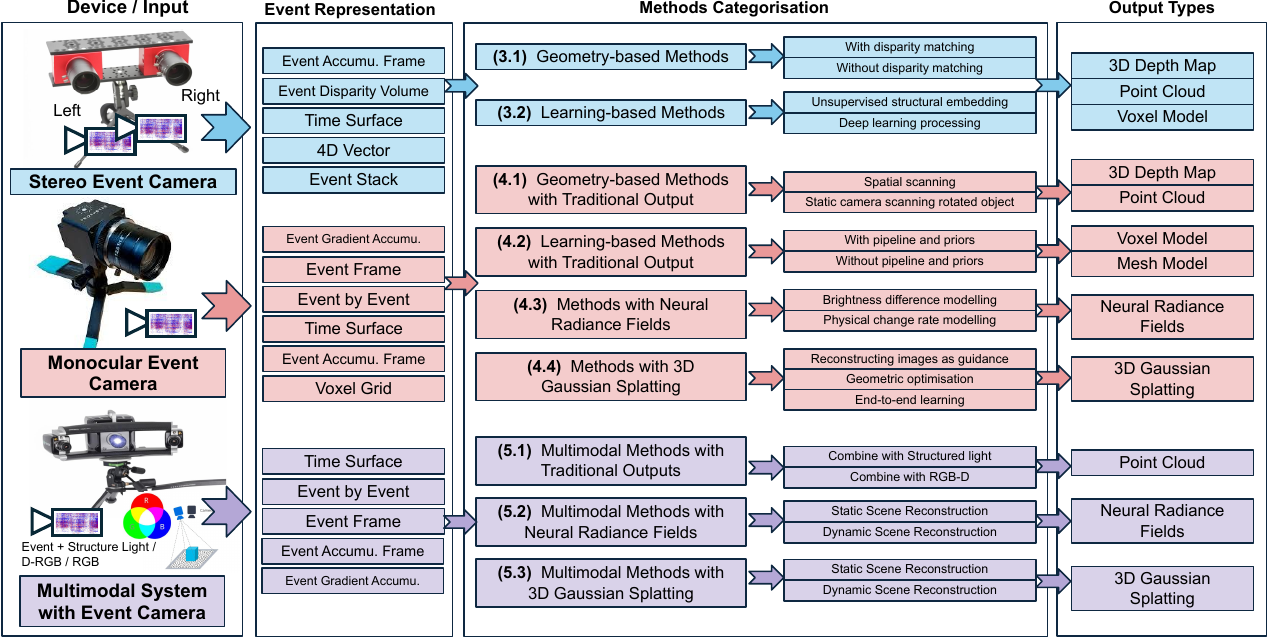}
    \caption{Categorisation of methods in the survey. This survey distinguishes stero, monocular, and multimodal event camera systems by their inputs, and further categorises methods by processing type and output type into geometric approaches, learning-based methods, and neural rendering frameworks based on NeRF and 3D Gaussian Splatting.}
    \label{fig:Categorisation}
\end{figure*}

To address this need, our survey aims to:

\begin{itemize}
\item Provide the first dedicated and comprehensive review specifically focusing on event-based 3D reconstruction.
\item Establish a coherent and structured categorisation of diverse event-driven 3D reconstruction methodologies.
\item Present a clear roadmap outlining technological progress and key milestones within the field.
\item Compile and summarise publicly available datasets related to event-driven 3D reconstruction.
\item Identify existing research gaps, suggest promising future directions, and discuss potential downstream applications of event-based 3D reconstruction.
\end{itemize}

This survey focuses on reconstruction methods that aim to recover modellable 3D representations of scene or object occupancy or rendering, specifically including recent neural rendering approaches such as Neural Radiance Fields and 3D Gaussian Splatting. However, important methods that focus on scene depth estimation are also considered.

As illustrated in Figure \ref{fig:roadmap}, Figure \ref{fig:timeline}, and Figure \ref{fig:Categorisation}, this paper systematically presents a hierarchical taxonomy and detailed chronological overview of event-based 3D reconstruction methods. In Section \ref{2}, we introduce the fundamental working principles of event cameras, discussing (\ref{2.1}) their differences from conventional cameras, (\ref{2.2}) the event generation mechanisms, and (\ref{2.3}) the typical event representations and output types commonly employed in reconstruction tasks. Sections \ref{3} through \ref{5} comprehensively review existing approaches categorised by camera setups and reconstruction methodologies. Specifically, Section \ref{3} focuses on methods utilising stereo event cameras, divided into (\ref{3.1}) geometry-based and (\ref{3.2}) learning-based methods. Section \ref{4} examines monocular event-camera techniques, further subdivided into (\ref{4.1}) geometry-based methods producing traditional outputs, (\ref{4.2}) learning-based methods producing traditional outputs, (\ref{4.3}) Neural Radiance Fields-based methods, and (\ref{4.4}) 3D Gaussian Splatting-based methods. Section \ref{5} addresses multimodal approaches integrating event cameras with complementary sensors, categorised into (\ref{5.1}) methods with traditional outputs, (\ref{5.2}) methods employing Neural Radiance Fields, and (\ref{5.3}) methods using 3D Gaussian Splatting. Subsequently, Section \ref{6} compiles publicly available datasets relevant for benchmarking event-driven 3D reconstruction techniques, and Section \ref{metrics} provides an overview of the most important and commonly used evaluation metrics in this field, categorized by type. Finally, Section \ref{7} summarises this review, highlighting current research gaps and outlining promising future directions to advance the field.

\section{Event Camera with 3D Reconstruction}  \label{2}
\subsection{Event Camera vs. Traditional Camera}  \label{2.1}

Event cameras have several distinctive characteristics that set them apart from traditional frame-based cameras. They offer microsecond-level temporal resolution \cite{lichtsteiner2008128}, enabling the capture of extremely fast motion without suffering from motion blur. With a dynamic range exceeding 120 dB \cite{lichtsteiner2008128}, event cameras can handle both extremely bright and very dark scenes, making them highly suitable for environments with challenging or rapidly changing lighting conditions.

\begin{figure}[t]
    \centering
    \includegraphics[width=\linewidth]{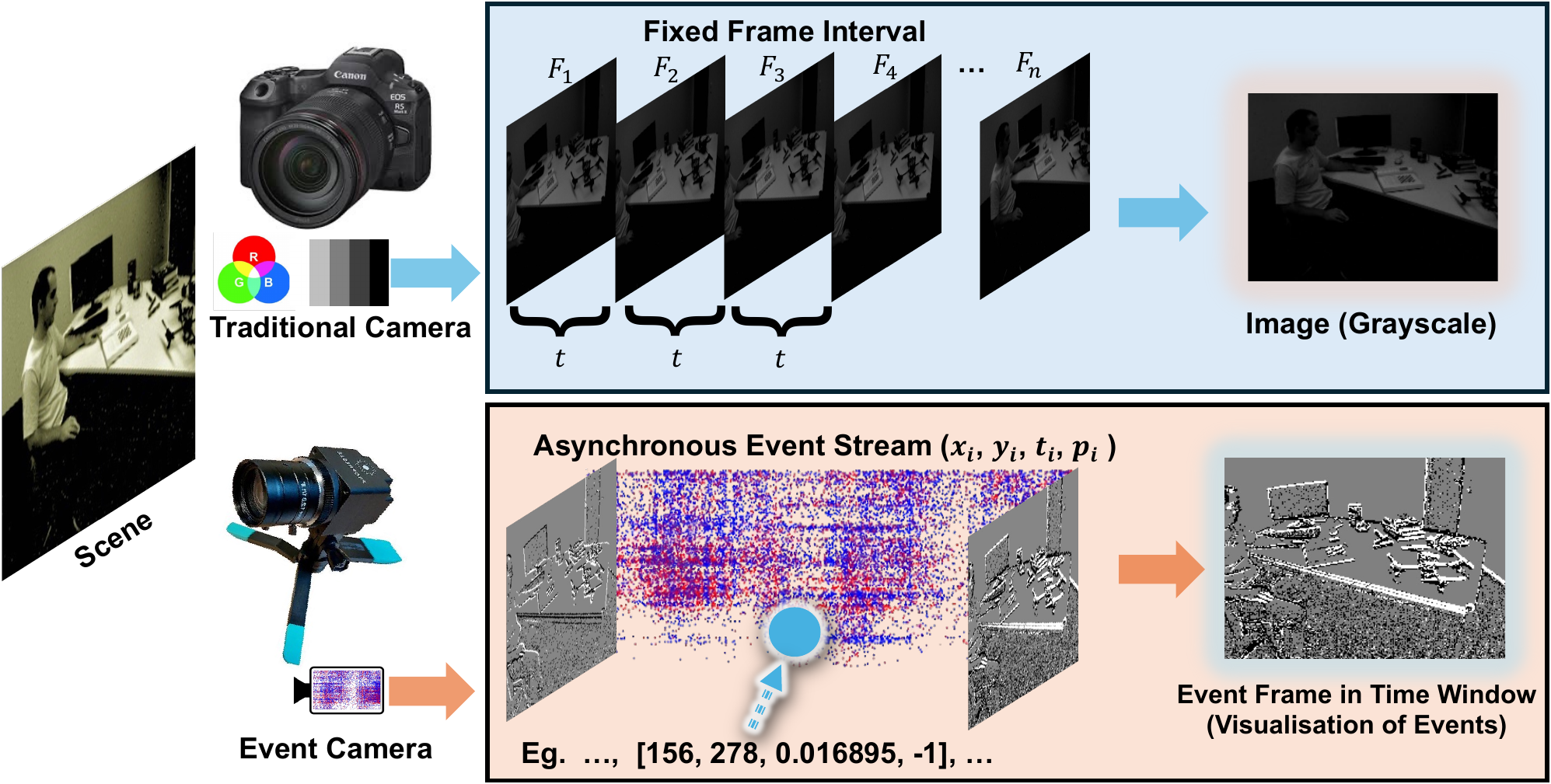}
    \caption{Event Camera vs. Traditional Camera. Traditional cameras output images at a fixed frame rate, whereas event cameras respond asynchronously to brightness changes in the scene, continuously generating a stream of events carrying spatial, temporal, and polarity information.}
    \label{fig:Event Camera vs. Traditional Camera}
\end{figure}

Referring to Figure \ref{fig:Event Camera vs. Traditional Camera} and Table \ref{tab:event_vs_traditional_compact}, unlike traditional cameras that capture full frames at fixed intervals, event cameras operate asynchronously by detecting changes in pixel intensity. Each pixel operates independently and triggers an event only when a significant change occurs \cite{lichtsteiner2008128}, resulting in a continuous stream of spatio-temporal events. This principle leads to significantly lower data rates, reduced power consumption, and minimal latency, which are highly desirable characteristics for real-time applications and edge computing scenarios.

%\begin{table*}[]
\begin{table*}
\centering
\caption{Compact Comparison between Event Cameras and Traditional Cameras}
\renewcommand{\arraystretch}{1.1}
\rowcolors{2}{gray!3}{gray!10}
%\footnotesize
\begin{tabularx}{\textwidth}{p{3.2cm} X X}
\toprule
%\rowcolor{gray!20}
\textbf{Aspect} & \textbf{Event Camera} & \textbf{Traditional Camera}\\ \hline
Acquisition & Asynchronous & Synchronous \\
Time Resolution & \textbf{1–10 $\mu$s/event }& 16–33 ms/frame (30–60 FPS) \\
Dynamic Range & \textbf{$>$120 dB} & 50–70 dB \\
Motion Blur & \textbf{Minimal} due to asynchronous capture & Significant at high-speed motion \\
Power Consumption & Typically \textbf{10–100 mW} & Typically 1–2 W \\
Data Redundancy & \textbf{Low} (only brightness change encoded) & High (entire frame captured) \\
Data Type & Spatiotemporal events: $(x, y, t, p)$ & RGB images: $H \times W \times 3$ matrices \\
Data Rate & \textbf{$\sim$0.1–2 MB/s} (scene dependent) & $\sim$30–300 MB/s (e.g., 1080p@30FPS) \\
Resolution & Low (e.g., 128×128, 240×180, 346×260) & High (e.g., 640×480, 1920×1080, 4K) \\
Latency & \textbf{$<$1 ms} (end-to-end) & 10–100 ms (sensor + processing) \\
Typical Use & High-speed, HDR, low-latency, robotics & General-purpose computer vision \\
Processing & Custom asynchronous pipelines or event representation & Compatible with standard CNNs \\
\bottomrule
\end{tabularx}
\label{tab:event_vs_traditional_compact}
\end{table*}

These unique characteristics make event cameras particularly well-suited for a variety of advanced computer vision tasks. Applications include, but are not limited to, high-speed object tracking~\cite{[35]}, low-latency corner detection~\cite{[JDJC]}, real-time object recognition~\cite{[36]}, and accurate depth estimation~\cite{[37]}. Event cameras have also shown great promise in video generation tasks~\cite{[SPSC], qu2025evanimate}, light field video enhancement~\cite{[LightField]}, and 3D reconstruction, where the combination of low latency and high temporal resolution contributes to precise and efficient scene understanding. With growing interest in neuromorphic vision and efficient visual sensing, event cameras are expected to play a key role in the future of robotics \cite{mahlknecht2022exploring}, AR/VR \cite{dong2024sevar}, autonomous driving \cite{maqueda2018event}, and beyond \cite{[R1]}.

\subsection{Event Triggering \& Event Representation}  \label{2.2}
A mathematical approach is very common to explain the triggering of events. When the event camera detects a brightness change at a pixel $k$, it generates event data containing event coordinates \( \mathbf{x}_k = (x_k, y_k) \), timestamp \( t_k\) , and the polarity \( p_k \). The brightness \( L(\mathbf{x}_k, t) = \log(I(\mathbf{x}_k, t)) \) is set as the pixel's log intensity. The brightness change threshold \( C \) usually varies by 10-15\% \cite{[EMVS]}. An event \( e_k = (\mathbf{x}_k, t_k, p_k) \) is triggered when the brightness change $\Delta L$ at a pixel $k$ exceeds $C$, which can be expressed as:
\begin{equation}
|\Delta L(\mathbf{x}_k, t_k)| = |L(\mathbf{x}_k, t_k) - L(\mathbf{x}_k, t_{k-1})| \geq |p_k \cdot C|,
\end{equation}
where \(t_{k-1}\) represents the timestamp of the last event at the same pixel. The polarity value \(p_k\) is determined as follows:
\begin{equation}
p_k =
\begin{cases} 
+1, & \text{if } \Delta L(\mathbf{x}_k, t_k) \geq C \\
-1, &\text{if } \Delta L(\mathbf{x}_k, t_k) \leq -C \\
\text{No event}, &\text{if } -C < \Delta L(\mathbf{x}_k, t_k) < C
\end{cases}
\end{equation}

When an event camera continuously captures events, it forms an event stream, which can be represented as a sequence of events ordered by timestamps:
\begin{equation}
\text{EventStream} = \{(t_k, x_k, y_k, p_k)\}_{k=1}^{N},
\label{eq:EventStream}
\end{equation}
where $N$ denotes the total number of recorded events.

In simple terms, an event camera generates asynchronous event data containing timestamps, coordinates, and polarity whenever a pixel sensor detects a brightness change that exceeds a certain threshold.

The event stream generated by an event camera is a continuous sequence of asynchronous and sparse data, where each event contains a timestamp, spatial coordinates, and polarity. This type of data is fundamentally different from conventional image frames. Due to its sparse and asynchronous nature, event data is difficult to directly extract features from and cannot be used as input to traditional deep learning models such as CNNs. Therefore, specialised preprocessing methods are typically required to convert event data into a more structured form, and this process is referred to as event representation. In addition to making the data more suitable for feature extraction, event representations help reduce data redundancy, improve computational efficiency, and enhance overall performance. Each representation method has its own strengths and limitations, and their applicability may vary depending on the task. Common event representation methods include the Event Frame, Time Surface, and Voxel Grid \cite{TimeSurface, EventFrame, VoxelGrid}. Event accumulate frame (EAF) and event gradient accumulation (EGA) are also very commonly used in current tasks with neural rendering output types \cite{[G3], [G13]}. Previous surveys \cite{[R1],[R3]} have already provided comprehensive overviews of these representations, and thus we do not elaborate further here. 

\subsection{Event-based 3D Reconstruction Types}  \label{2.3}

There are many tasks in 3D vision, but those considered as complete 3D reconstruction tasks typically require the final output to faithfully recover the spatial geometry of a scene or object and to be represented in an explicit and well-defined 3D format \cite{geiger2011stereoscan}. Such tasks go beyond estimating depth, disparity, or motion, which focus on reconstructing and modelling the full 3D structure of the environment. Specifically, the output should be a point cloud, mesh, voxel representation, an implicit 3D representation (such as SDF \cite{[SDF]}, NeRF \cite{[Nerf]}, or Occupancy Networks \cite{[occupancy]}), etc., which can be directly used in downstream applications like rendering, path planning, robotic navigation, or virtual reality.

\begin{figure}[t]
    \centering
    \includegraphics[width=\linewidth]{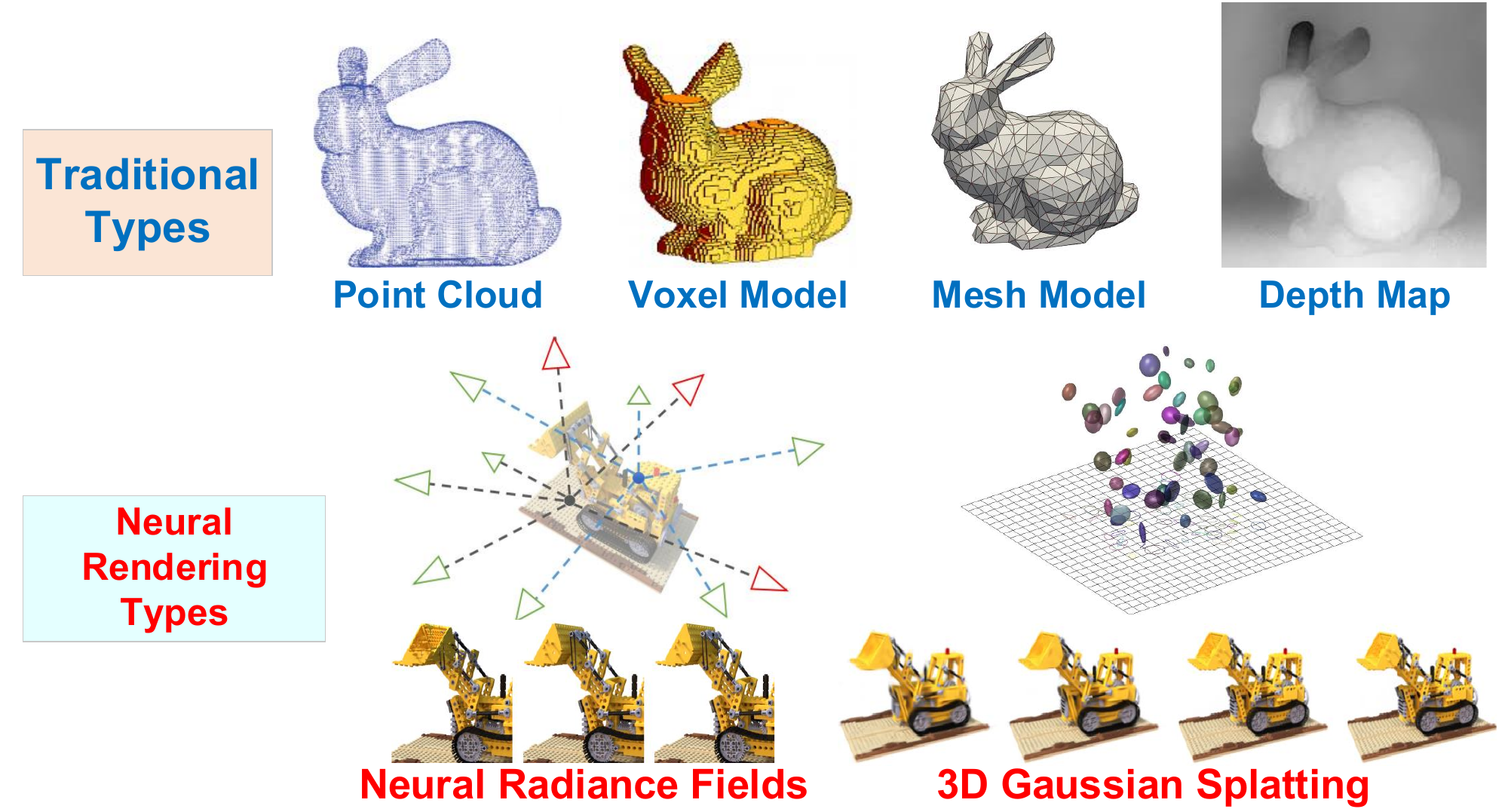}
    \caption{Different types of event-based 3D reconstruction results.}
    \label{fig:event_3d_types}
\end{figure}

In this survey, we focus exclusively on works that aim to reconstruct 3D structures as their primary objective and output a modelable 3D representation. We summarise the common output types in event-based 3D reconstruction tasks. The visualisations are shown in Figure \ref{fig:event_3d_types}. First, here are four traditional types of 3D reconstruction results:

\begin{itemize}
    \item \textbf{Point Cloud:} The point cloud is one of the most straightforward 3D representations. A point cloud consists of a set of 3D coordinates \((x, y, z)\) \cite{PointCloud}, with each point potentially associated with additional attributes such as colour, normals, or timestamps \cite{deng2024multi, guo2021pct}. Point clouds can be obtained by back-projecting disparities estimated via stereo matching or depth estimation.
    \item \textbf{Voxel Model:} Voxels discretise 3D space into regular cubic cells, where each voxel stores occupancy information or a probability value \cite{Voxel}. This representation is well-suited for volumetric modelling and is often used as input or supervision for neural networks.
    \item \textbf{Mesh Model:} A mesh consists of vertices, edges, and faces that define the surface of a 3D object \cite{Mesh}. It is typically generated from point clouds or voxel representations through surface reconstruction techniques. Although the sparse nature of event data presents challenges for mesh generation, recent methods have demonstrated that surface reconstruction from event streams is feasible.
    \item \textbf{Depth Map:} The depth map is a 2D image where each pixel encodes the depth value of the corresponding point in the scene \cite{saxena2008make3d}. Since a depth map is a 2D representation, it is less likely to be considered a complete 3D representation compared to the former three. However, with known camera intrinsics, a depth map can be easily converted into a point cloud, enabling more advanced surface modelling. Moreover, real-time depth estimation with event cameras is a topic closely related to, but distinct from, 3D reconstruction. Therefore, this survey includes some methods with 3D depth map output, and the following sections may cover crucial depth estimation works that produce 3D depth maps.
\end{itemize}

While traditional outputs are suitable for explicit geometry modelling in tasks such as SLAM \cite{laidlow2019deepfusion} and robotics \cite{banerjee2018robotic}, they often fall short in producing photorealistic renderings or modelling complex lighting and material properties. To address this, recent advances in neural rendering have introduced new 3D representations that are capable of modelling scenes in a continuous and differentiable manner \cite{yang2023jnerf}. These representations have enabled breakthroughs in novel view synthesis and high-fidelity scene reconstruction:

\begin{figure*}[t]
    \centering
    \includegraphics[width=\linewidth]{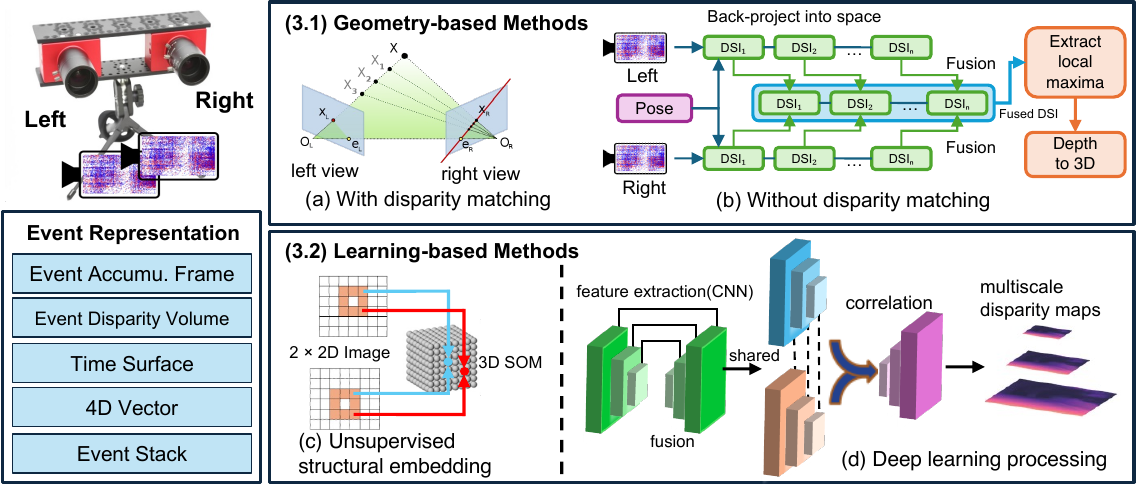}
    \caption{Overview of methods with stereo event cameras. (a) Schematic of disparity matching. (b) Refocused DSI Fusion to replace disparity matching module, from Ghosh et al. (c) Self-Organising Maps from Steffen et al. (d) UNet-ResNet (CNN) from Nam et al.}
    \label{fig:Stereo Event Methods}
\end{figure*}

\begin{itemize}
    \item \textbf{Neural Radiance Fields:} Neural Radiance Fields (NeRF) is a neural network-based method for representing 3D scenes, proposed by Mildenhall et al. in 2020 \cite{[Nerf]}. It learns a continuous and differentiable 3D radiance field from multi-view 2D images and synthesises novel views \cite{qu2024nerf}, modelling the complex geometries and lighting effects \cite{zhuang2025simple, liu2025sem}. Since 2023, NeRF-based methods have been adapted for event-based 3D reconstruction \cite{[N3]}. A 3D scene can be reconstructed by applying differentiable volume rendering and extracting surfaces via iso-surface techniques. A NeRF models a continuous volumetric scene by training a multilayer perceptron (MLP) to learn a mapping from a 3D spatial location \(\mathbf{x} \in \mathbf{R}^3\) and viewing direction \(\mathbf{d}\) to the corresponding colour and volume density:
\begin{equation}
    f(\mathbf{x}, \mathbf{d}) \rightarrow (\mathbf{c}, \sigma),
\end{equation}
where \(\mathbf{c}\) denotes the emitted colour and \(\sigma\) is the volume density at the point \(\mathbf{x}\).

\item \textbf{3D Gaussian Splatting (3DGS):} 3D Gaussian Splatting, proposed by Kerbl et al. \cite{[Gaussian]}, represents a volumetric primitive in 3D space with attributes such as position, shape, orientation, and colour. In computer graphics, it serves as an explicit 3D representation that enables efficient differentiable rendering \cite{advance3DGS}. It provides an efficient solution that lies between traditional explicit point clouds and implicit volumetric rendering.

Each Gaussian is defined by a 3D position \(\mu_i\), a covariance matrix \(\Sigma_i\), and additional attributes such as colour and opacity. The contribution of the \(i\)-th Gaussian to a spatial point \(\mathbf{p}\) is defined by:
\begin{equation}
\hspace*{-0.25cm}
f_i(\mathbf{p}) = \sigma(\alpha_i) \exp\left( 
 -\frac{1}{2} (\mathbf{p} - \mu_i)^T \Sigma_i^{-1} (\mathbf{p} - \mu_i) 
\right)
\end{equation}
where \(\sigma(\alpha_i)\) represents the activated opacity, \(\mu_i\) is the center of the Gaussian, and \(\Sigma_i\) controls its spatial spread and orientation. Each Gaussian can be seen as a soft volumetric primitive, whose projection onto the image plane contributes to the final image through \(\alpha\)-blending.
\end{itemize}

\section{Methods with Stereo Event Cameras}  \label{3}
Stereo event cameras typically refer to two or more rigidly mounted event cameras. A stereo vision system composed of stereo event cameras is capable of asynchronously capturing event streams from the left and right viewpoints of a scene, enabling event-driven stereo matching and 3D scene reconstruction \cite{zhu2018realtime}. In general, the goal of stereo matching is to identify the most probable pairs of left and right events in both space and time, in order to compute disparity and subsequently estimate the depth structure of the scene. 

Event-driven 3D-related tasks, such as stereo vision, were initially pioneered using stereo event cameras \cite{mahowald1994vlsi}, and there is more research solving stereo vision \cite{[S3],[S100]}. However, these works do not achieve a fully modelled 3D reconstruction as the final output, and therefore are not included in the survey.

Based on the data processing pipeline, 3D reconstruction methods using only stereo event cameras can be categorised into: (\ref{3.1}) geometry-based methods and (\ref{3.2}) learning-based methods. Table \ref{tab:stereo_event_methods} and Figure \ref{fig:Stereo Event Methods} provide an overview of these methods. Figure \ref{fig:depthmap_comparison} provides a visualisation of some of these methods producing depth map output.

\setlength{\fboxsep}{0pt}
\setlength{\fboxrule}{0.4pt}
\newcommand{\imgbox}[1]{\fbox{\includegraphics[width=0.1\textwidth]{#1}}}
\begin{figure}[t]
\centering
{\small
\setlength{\tabcolsep}{1pt}
\resizebox{\linewidth}{!}{ 
\begin{tabular}{
>{\centering\arraybackslash}m{0.1\textwidth}
>{\centering\arraybackslash}m{0.1\textwidth}
>{\centering\arraybackslash}m{0.1\textwidth}
>{\centering\arraybackslash}m{0.1\textwidth}
>{\centering\arraybackslash}m{0.1\textwidth}}
\textbf{Scene} 
& \textbf{(a)} 
& \textbf{(b)} 
& \textbf{(c)} 
& \textbf{(d)} \\

\imgbox{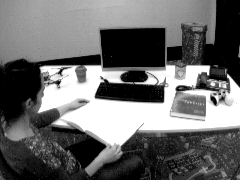} & \imgbox{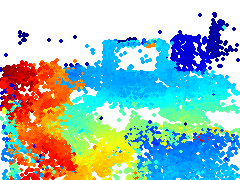} & \imgbox{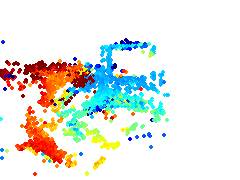} & \imgbox{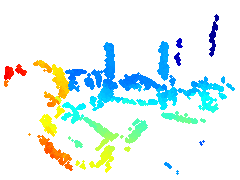} & \imgbox{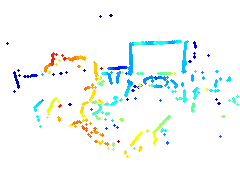} \\[-1.5pt]
\imgbox{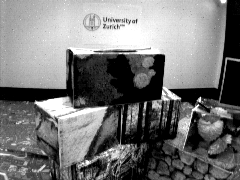} & \imgbox{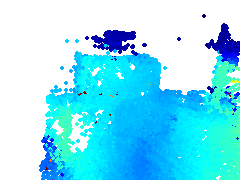} & \imgbox{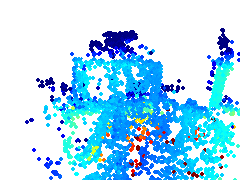} & \imgbox{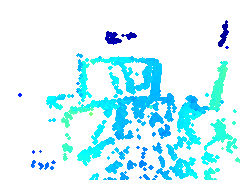} & \imgbox{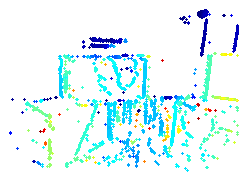} \\[-1.5pt]
\imgbox{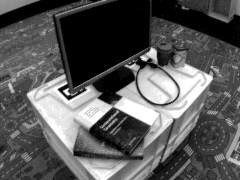} & \imgbox{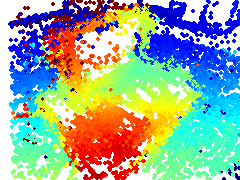} & \imgbox{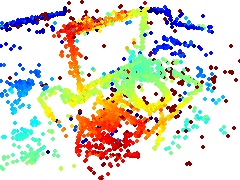} & \imgbox{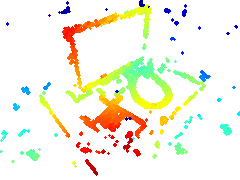} & \imgbox{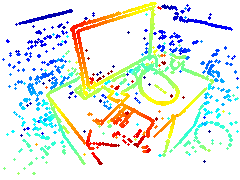} \\[-1.5pt]
\imgbox{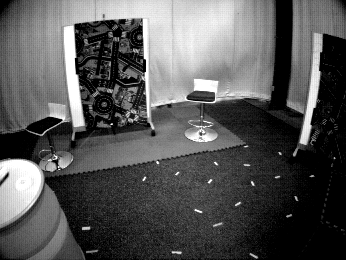} & \imgbox{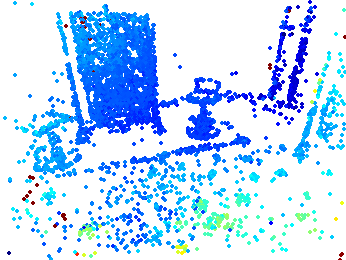} & \imgbox{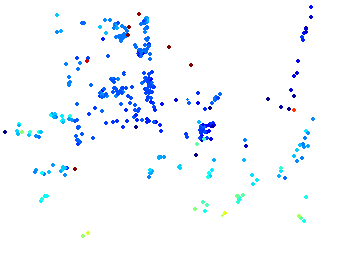} & \imgbox{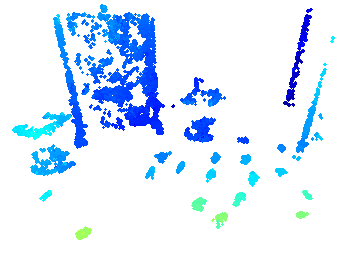} & \imgbox{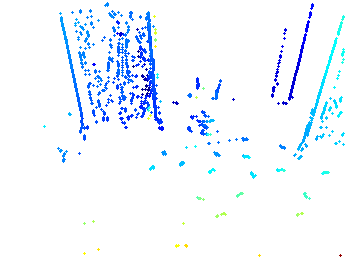} \\

\end{tabular}
}
}
\caption{
Comparison of depth maps generated from methods using a stereo event camera, adapted from Ghosh et al., under CC BY 4.0, @ Wiley \& Sons 2022. (a) SGM method (RGB-based) is adapted to event data by Ghosh et al. (b) GTS method from Leng et al. (c) ESVO method from Zhou et al. (d) Method from Ghosh et al.
}
\label{fig:depthmap_comparison}
\end{figure}

\begin{table*}[th!]
    \caption{Methods with \textbf{stereo} event cameras}
    \centering
    \resizebox{\textwidth}{!}{%
    \rowcolors{2}{cyan!5}{cyan!15}
    \begin{tabular}{lcccccc}
        \toprule
        \textbf{Author} & \textbf{Year} & \textbf{Priors} & \textbf{Event Rep.} & \textbf{Real-time} & \textbf{Output (Dense?)} & \textbf{Dataset} \\
        \hline
        Schraml et al. \cite{[S9]} & 2015 & Trajectory & EAF & \ding{51} & 3D depth map (\ding{55}) & Self-collected \cite{[S9]} \\
        Zhu et al. \cite{[S0]} & 2018 & Velocity & Event disparity volume & \ding{51} & 3D depth map (\ding{51}) & MVSEC \cite{MVSEC} \\
        Zhou et al. \cite{[S2]} & 2018 & Pose & Time surface & \ding{55} & 3D depth map (\ding{55}) & MVSEC \cite{MVSEC} \\
        Leng et al. \cite{[S4]} & 2018 & Pose & Time surface & \ding{51} & Point cloud (\ding{55}) & Self-collected \cite{[S4]} \\
        Zhu et al. \cite{[XS2]} & 2019 & Pose & Event disparity volume & \ding{55} & 3D depth map (\ding{55}) & Self-collected \cite{[S5]} \\
        Steffen et al. \cite{[S5]} & 2019 & - & 4D vector & \ding{55} & Voxel (\ding{55}) & MVSEC \cite{MVSEC} \\
        Zhou et al. \cite{[S6]} & 2021 & Pose & Time surface & \ding{51} & Point cloud (\ding{55}) & MVSEC \cite{MVSEC} \\
        Nam et al. \cite{[S7]} & 2022 & Pose & Event Stack & \ding{51} & 3D depth map (\ding{51}) & DSEC \cite{DSEC2}\\
        Ghosh et al. \cite{[S8]} & 2022 & Pose & Event disparity volume & \ding{51} & 3D depth map (\ding{55}) & MVSEC \cite{MVSEC}, DSEC \cite{DSEC2}, etc \\
        Ghosh et al. \cite{[XS3]} & 2024 & Pose & Event disparity volume & \ding{51} & 3D depth map (\ding{55}) & MVSEC \cite{MVSEC}, DSEC \cite{DSEC2}, etc \\
        Freitag et al. \cite{[XS1]} & 2025 & Pose & Event Stack & \ding{55} & Point cloud (\ding{51}) & Self-collected \cite{[XS1-data]} \\
        Hitzges et al. \cite{[XS4]} & 2025 & Pose & Event disparity volume & \ding{55} & 3D depth map (\ding{55}) & MVSEC \cite{MVSEC}, DSEC \cite{DSEC2} \\
       \bottomrule
    \end{tabular}
    }
    \label{tab:stereo_event_methods}
\end{table*}

\subsection{Geometry-based Methods} \label{3.1}
Geometry-based methods using stereo event cameras typically aim to recover depth by leveraging the known stereo baseline and spatial-temporal event correspondences. While many early approaches compute disparity through explicit matching across viewpoints, others bypass this step by directly optimising consistency measures in the temporal or geometric domain. Based on whether explicit disparity computation is required, we categorise these methods into the following two types:

\subsubsection{With disparity matching}
Many earlier approaches perform depth mapping reconstruction by computing disparity between events observed at the same timestamp across different viewpoints, followed by geometry-based multi-view stereo estimation to achieve real-time 3D depth reconstruction. In 2015, Schraml et al. \cite{[S9]} introduced a novel stereo matching method for a rotating panoramic stereo event system. Their approach defines a cost metric based on event distribution similarity between left and right views and computes disparity through dynamic programming on sparse event maps. The resulting disparities are then used to reconstruct real-world depth in 360° panoramic views. In 2018, Zhu et al. \cite{[S0]} proposed a method that synchronises events in time using known camera velocities, constructs a dense event disparity volume, and performs real-time sliding window matching, introducing a novel matching cost function combining ambiguity and similarity. Simultaneously, Leng et al. \cite{[S4]} reformulated event matching as a time-based stereo vision problem, where disparity estimation is achieved by detecting the temporal coincidence of events across epipolar lines. Their method introduces a generalised spiking neuron model that integrates spatio-temporal event information with temporal decay, allowing robust matching under varying motion and illumination conditions. A voting scheme is then used to infer depth from the accumulated neural activations, enabling dense 3D reconstruction from pure event streams. In 2025, Freitag et al. \cite{[XS1]} proposed a stereo reconstruction method that performs disparity matching by leveraging temporally coinciding events between two event cameras. By triangulating matched event pairs, their system reconstructs high-precision point clouds.

\subsubsection{Without disparity matching}
Disparity matching can also be bypassed altogether by directly optimising geometric or temporal consistency across event streams. In 2018, Zhou et al. \cite{[S2]} proposed a novel forward-projection-based depth estimation method, which avoids the need for explicit disparity computation. Instead of finding correspondences between stereo events, their method optimises a temporal consistency energy by projecting candidate depth hypotheses into both stereo time surfaces and measuring their alignment. A coarse-to-fine search strategy is used to obtain semi-dense depth maps in real-time, enabling robust depth reconstruction under challenging conditions. Building upon this idea, Zhou et al. \cite{[S6]} extended the concept to a stereo event-based visual odometry framework. By jointly optimising over spatial and temporal consistency across stereo time surfaces, they estimate not only per-frame depth but also the stereo camera motion. The system runs in real-time and outputs semi-dense depth maps aligned with visual odometry estimates, making it well-suited for 3D reconstruction in dynamic environments. Later in 2022, Ghosh et al. \cite{[S8]} proposed a multi-event-camera fusion framework that leverages Disparity Space Images (DSIs) \cite{DSI} to accumulate event ray densities from multiple viewpoints. Instead of requiring event correspondences, their method refocuses events in DSI space and performs probabilistic fusion to obtain high-quality depth maps. They further introduce an outlier rejection strategy to enhance robustness and demonstrate generalisation across multiple datasets without fine-tuning. Building upon this idea, in 2024, Ghosh et al. \cite{[XS3]} proposed ES-PTAM, which combines a mapping module based on ray density fusion and a tracking module using edge-map alignment, both operating on pure event data. By bypassing disparity matching and processing stereo events in a correspondence-free manner via geometric ray density fusion, the system produces accurate semi-dense depth maps and camera poses in real time. It is a leading stereo visual odometry method that exclusively uses events, without any auxiliary inputs.

\subsection{Learning-based Methods} \label{3.2}
Learning-based methods leverage neural networks to estimate 3D structure from stereo event data. Based on the learning strategy, we categorise them into the following two types:

\subsubsection{Unsupervised structural embedding}
Additionally, one research explores unsupervised clustering for self-organising structural modelling. In 2019, Steffen et al. \cite{[S5]} proposed an unsupervised reconstruction framework using Self-Organising Maps (SOMs) \cite{[S5-13]}. By embedding stereo events from multiple viewpoints into a shared latent space, their method performs sparse voxel-based 3D reconstruction without requiring calibration or supervision, offering a lightweight and biologically inspired approach to structural modelling.

\subsubsection{Deep learning processing}
More recent methods leverage deep neural networks to learn stereo disparity estimation and event representations. In 2019, Zhu et al. \cite{[XS2]} proposed an unsupervised deep learning framework, which leverages neural networks to predict depth and motion from stereo event data. They introduce a discretized event volume representation to preserve spatio-temporal structure and apply a motion compensation loss that reduces event blur. In 2022, Nam et al. \cite{[S7]} introduced a deep stereo framework that combines multi-density event stacking with attention-guided encoding via a UNet-ResNet \cite{[S7-35],[S7-14]} backbone. Their method further incorporates future event prediction during training through a distillation loss \cite{[S7-15],[S7-38]}, enabling real-time, dense depth estimation with high accuracy on the DSEC dataset \cite{[S7-13]}. In 2025, Hitzges et al. \cite{[XS4]} proposed DERD-Net, which estimates depth from event-based ray densities by processing DSIs derived from multi-view event data and known camera poses. By extracting local sub-volumes (Sub-DSIs) and combining 3D convolutions with recurrent units, their method enables efficient, high-resolution depth prediction.

\begin{figure*}[th]
    \centering
    \includegraphics[width=\linewidth]{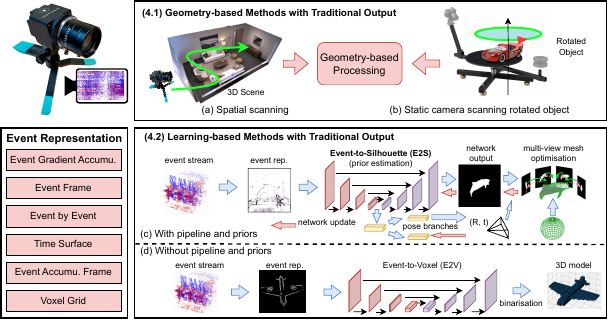}
    \caption{Overview of methods with monocular event camera that generate traditional output. (a) Schematic of spatial scanning in a 3D scene. (b) Schematic of static event camera scanning rotated object on a platform. (c) The E3D pipeline that uses E2S to estimate physical prior - Silhouette, to assist 3D mesh reconstruction. (d) The end-to-end structure that directly generate 3D model from Chen et al.}
    \label{fig:Monocular1}
\end{figure*}

\section{Methods with Monocular Event Cameras}  \label{4}

Unlike stereo event cameras, which can directly observe scene disparities from two synchronised viewpoints, monocular event cameras have only a single viewpoint and thus cannot infer depth from direct geometric correspondences. As a result, 3D reconstruction using monocular event cameras often requires or estimates additional prior information or assumptions related to known motion, scene rigidity, or temporal consistency, to compensate for the lack of stereo disparity cues.

Despite these limitations, monocular setups are more lightweight, power-efficient, and easier to deploy and move, making them suitable for embedded systems and mobile equipment. Consequently, monocular event-based 3D reconstruction has attracted increasing research interest and covers a broad range of tasks and 3D representation strategies.

Existing monocular approaches can be broadly divided into four categories based on their underlying modelling techniques and output formats: (\ref{4.1}) geometry-based methods with traditional outputs, (\ref{4.2}) learning-based methods with traditional outputs, (\ref{4.3}) methods with Neural Radiance Fields, and (\ref{4.4}) methods with 3D Gaussian Splatting. The first two categories aim to reconstruct explicit 3D outputs such as depth maps, point clouds, meshes, or voxels, while the latter two leverage volumetric rendering paradigms to generate photorealistic and continuous scene representations from asynchronous event streams.

Additionally, some methods leverage RGB Bayer event streams captured by colour event sensors like Colour DAVIS (e.g., DAVIS 346c) to support colour-consistent NeRF and Gaussian rendering \cite{[N2], [G5], [G11]}. However, since the colour is encoded within the same event modality, these approaches are still considered monocular, not multimodal.

\subsection{Geometry-based Methods with Traditional Output}
\label{4.1}

Geometry-based methods typically achieve semi-dense or sparse 3D reconstruction in real-time, relying on spatial-temporal event aggregation and geometric priors. Since monocular event cameras cannot directly observe disparities, these methods often require known or estimated motion to compensate for the view angles. Figure \ref{fig:Monocular1} and Table \ref{tab:mono_geom} provide an overview of these methods. Based on the data acquisition strategy, we categorise these methods into the following two types:

% \noindent\textbf{Spatial scanning: }
\subsubsection{Spatial scanning}
Many methods rely on spatial scanning, where the event camera must move through space to accumulate multi-view observations. By capturing asynchronous events from different viewpoints, these methods infer scene structure through motion-induced parallax. In 2016, Kim et al. \cite{[M1]} proposed an approach utilising three interleaved probabilistic filters to estimate camera trajectory, scene log-intensity gradient, and inverse depth. In 2016, Rebecq et al. \cite{[EMVSconf]} proposed EMVS, employing event space-sweep and ray density analysis to directly generate a semi-dense 3D depth map, without frame-level data association. Later, EMVS was extended into a full SLAM framework as its mapping module, where Rebecq et al. \cite{[M2]} introduced EVO (Event-based Visual Odometry), which integrates event projection, time-surface alignment, and DSI-based depth fusion to jointly estimate camera trajectory and reconstruct 3D geometry in real time. A recent study has improved geometry-based methods to achieve dense reconstruction. In 2024, Guan et al. \cite{[M3]} proposed EVI-SAM, a tightly coupled event-image-IMU SLAM framework. It achieves real-time dense 3D reconstruction on a standard CPU, integrating event-based 2D-2D alignment, image-guided depth interpolation, and TSDF fusion \cite{[M3-58]}.

\subsubsection{Static camera scanning rotated object}
A unique innovation enables reconstruction when the event camera is stationary, while the object rotates. In 2024, Elms et al. \cite{[M4]} proposed eSfO, which performs 3D reconstruction through event corner tracking and factor graph optimisation \cite{[M4-12]}, but it only performs non-real-time sparse point cloud reconstruction.

\begin{table*}[th!]
    \centering
    \caption{\textbf{Monocular} event camera: \textbf{Geometry-based} methods with traditional outputs} % 
    \resizebox{0.95\textwidth}{!}{%
    \rowcolors{2}{red!3}{red!10}
    \begin{tabular}{ccccccc}
    \toprule
        \textbf{Author} & \textbf{Year} & \textbf{Priors} & \textbf{Event Rep.} & \textbf{Real-time} & \textbf{Output (Dense?)} & \textbf{Dataset} \\
    \hline
        Kim et al. \cite{[M1]}&2016&Trajectory&EGA&\ding{51}&3D depth map (\ding{55})& Self-collected \cite{[M1]}\\
        Rebecq et al. \cite{[EMVSconf]}&2016&Trajectory&Event-by-event&\ding{51}&3D depth map (\ding{55})& Self-collected \cite{[EMVS]}\\
        Rebecq et al. \cite{[M2]}&2016&Pose&Event frame&\ding{51}&3D depth map (\ding{55})& Self-collected \cite{[M2]}\\
        Guan et al. \cite{[M3]}&2024&Trajectory&Time surface&\ding{51}&3D depth map (\ding{51})& DAVIS240C \cite{DAVIS240C}\\
        Elms et al. \cite{[M4]}&2024&Trajectory&Time surface&\ding{55}&Point cloud (\ding{55})& TOPSPIN \cite{[M4]}\\
    \bottomrule
    \end{tabular}
    }
    \label{tab:mono_geom}
\end{table*}

\begin{table*}[th!]
    \centering
     \caption{\textbf{Monocular} event camera: \textbf{Learning}-based methods with traditional outputs}
    \resizebox{\textwidth}{!}{%
    \rowcolors{2}{red!3}{red!10}
    \begin{tabular}{cccccccc}
    \toprule
        \textbf{Author} & \textbf{Year} & \textbf{Priors} & \textbf{Pipeline} & \textbf{Event Rep.} & \textbf{Model} & \textbf{Output (Dense?)} & \textbf{Dataset}\\
        \hline
        Baudron et al. \cite{[MD1]} & 2020 & Silhouette & \ding{51} & EAF & E2S(CNN) & Mesh (\ding{51}) & ShapeNet \cite{[MD1-D]} \\
        Xiao et al. \cite{[MD2]} & 2022 & Pose & \ding{51} & Voxel grid & E2VID(RNN-CNN) & Mesh (\ding{51}) & ESIM \cite{[esim]}\\
        Wang et al. \cite{[MD5]} & 2023 & Contour, Trajec. & \ding{51} & Voxel grid & Evac3d(CNN) & Mesh (\ding{51}) & MOEC-3D \cite{[MD5]} \\
        Chen et al. \cite{[Dense]} & 2023 & - & \ding{55} & Event frame & E2V(CNN)& Voxel (\ding{51}) & SynthEVox3D \cite{[Dense]}\\
        Xu et al.  \cite{[Towards]} & 2025 & - & \ding{55} & Event frame & E2V(CNN)& Voxel (\ding{51}) & SynthEVox3D \cite{[Dense]}\\
        \bottomrule
    \end{tabular}
    \label{tab:mono_deep_learning}
    }
\end{table*}

\subsection{Learning-Based Methods with Traditional Output}
\label{4.2}

Deep learning-based methods typically produce non-real-time dense reconstruction. However, traditional RGB image feature extraction techniques cannot be directly applied to raw event data, and image-based event representation is often required as a preprocessing \cite{[N7]}. Table \ref{tab:mono_deep_learning} and Figure \ref{fig:Monocular1} provide an overview of these methods. Based on the reliance on structured pipelines and external priors, we categorise these methods into the following two types:

\subsubsection{With pipeline and priors}
Many studies have established an event-to-3D pipeline, a structured and modular event processing framework \cite{[MD1], [MD2], [N7], [M2]}, including feature extraction, matching, and 3D computation. The extraction and estimation of priors are also essential. In 2020, Baudron et al. \cite{[MD1]} proposed E3D, the first dense 3D shape reconstruction method based on monocular event cameras. It employs the E2S neural network to estimate silhouettes and leverages PyTorch3D \cite{[MD1-29]} for 3D mesh optimisation, achieving high-quality multi-view 3D reconstruction trained on ShapeNet \cite{[MD1-5]}. In 2022, Xiao et al. \cite{[MD2]} proposed a pipeline using the E2VID deep learning method \cite{[T4]} to process continuous event streams and generate normalised intensity image sequences. They then employed SfM \cite{[MD2-17]} to estimate intrinsic and extrinsic parameters for sparse point clouds and used MVS \cite{[MD2-28],[MD2-29],[MD2-30],[MD2-31]} for dense mesh reconstruction. In 2023, Wang et al. \cite{[MD5]} proposed EvAC3D, which uses CNN to predict Apparent Contour Events (ACE), combined with Continuous Volume Carving and Global Mesh Optimisation, to achieve dense 3D shape reconstruction with known camera trajectories.

\subsubsection{Without pipeline and priors}
However, recent methods aim to eliminate the pipeline and priors. In 2023, Chen et al. \cite{[Dense]} proposed E2V, which employs a modified ResNet-152 and a U-Net 3D decoder to directly predict dense 3D voxel grids from monocular event frames, achieving event-based 3D reconstruction without external priors. In 2025, Xu et al. \cite{[Towards]} extended E2V by introducing a novel event representation, Sobel Event Frame, and an optimal binarisation strategy for event-based 3D reconstruction. By enhancing E2V with Efficient Channel Attention \cite{[Towards-21], guo2022attention}, their method significantly improved reconstruction quality.

\subsection{Methods with Neural Radiance Fields} \label{4.3}
\begin{figure*}[ht!]
    \centering
    \includegraphics[width=\linewidth]{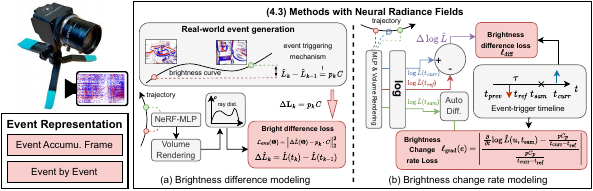}
    \caption{Overview of methods with monocular event cameras that produce Neural Radiance Fields. (a) The method structure mainly modelling brightness difference, $\Delta \hat{L}_k$, from Klenk et al. (b) The method structure that also models brightness change rate, $\frac{\partial}{\partial t} \log \hat{L}(\mathbf{u}, t_{\text{sam}})$, from Low et al.}
    \label{fig:Monocular2}
\end{figure*}
\begin{table*}[t]
    \centering
    \caption{\textbf{Monocular} event camera: \textbf{NeRF}-based Methods}
    \resizebox{\textwidth}{!}{%
    \rowcolors{2}{red!3}{red!10}
    \begin{tabular}{ccccccc}
        \toprule
        \textbf{Author} & \textbf{Model} & \textbf{Yr-Mo} & \textbf{Inputs} & \textbf{Event Rep.} & \textbf{Colourful} & \textbf{Dataset} \\
        \hline
        Klenk et al. \cite{[N3]} & E-NeRF & 2023-01 & Event stream & EAF & \ding{55} & ESIM \cite{[esim]}\\
        Hwang et al. \cite{[N1]} & Ev-NeRF & 2023-03 & Event stream & EAF & \ding{55} & IJRR \cite{IJRR}, HQF \cite{HQF} \\
        Rudnev et al. \cite{[N2]} & EventNeRF & 2023-06 & RGB bayer event stream & EAF & \ding{51} & Self-collected \cite{[N2]} \\
        Low et al. \cite{[N13]}& Robust e-NeRF& 2023-10 & Event stream & Event-by-event & \ding{51} & TUM-VIE \cite{TUM-VIE}\\
        Bhattacharya et al. \cite{[N4]} & EvDNeRF & 2024-01 & Event stream & EAF & \ding{55} & Self-collected \cite{[N4]} \\
        Wang et al. \cite{[N7]} & NeRF(Enhanced) & 2024-05 & Event stream & EAF & \ding{55} & PAEv3D \cite{[N7]}\\
        Low et al. \cite{[XN2]} & Deblur e-NeRF & 2024-09 & Event stream & Event-by-event & \ding{51} & EDS \cite{EDS}\\
        Feng et al. \cite{[N8]} & AE-NeRF & 2025-04 & Event stream & Event-by-event & \ding{51} & TUM-VIE \cite{TUM-VIE}\\
        Wang et al. \cite{[XN4]} & SaENeRF & 2025-04 & Event stream & EAF & \ding{51} & Rudnev et al. \cite{[N2]} \\
        \bottomrule
    \end{tabular}
    }
    \label{tab:mono_nerf}
\end{table*}
% \begin{figure}[h]
\begin{figure}[t]
    \centering
    \includegraphics[width=\linewidth]{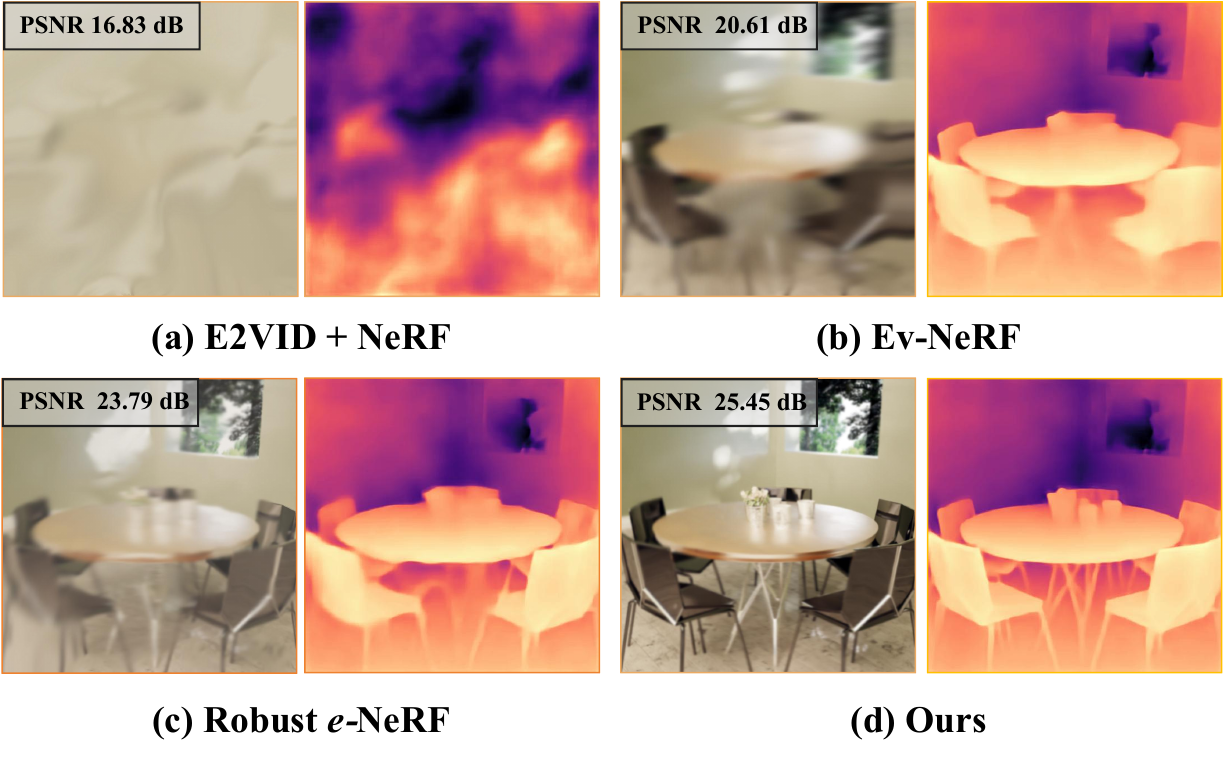}
    \caption{Visual comparison of several NeRF-based methods using monocular event cameras, adapted from Feng et al., under CC BY-NC-SA 4.0. (a) Result from the monocular deep learning method E2VID combined with NeRF. (b) Ev-NeRF method. (c) Robust e-NeRF method. (d) AE-NeRF method.}
    \label{fig:N8}
\end{figure}

Compared to geometry-based or deep learning-based methods, NeRF-based methods avoid explicit feature extraction or geometry construction pipelines. Instead, they rely on end-to-end optimisation guided by photometric or event-based supervision to recover implicit scene structure. NeRF-based methods typically use the Event Accumulate Frame as the main input representation of event streams and are generally designed for grayscale 3D reconstruction. Particularly, some methods achieve colour 3D reconstruction using only the event stream as input. Low et al. \cite{[N13],[XN2]} restore colour using gamma correction. Feng et al. \cite{[N8]} adopt a learning-based colour correction method. Wang et al. \cite{[XN4]} leverage NeRF’s volume rendering and colour modelling to perform self-supervised colour 3D reconstruction without RGB image supervision. Table \ref{tab:mono_nerf} and Figure \ref{fig:Monocular2} provide an overview of these methods, and Figure \ref{fig:N8} provides a visualisation example of some methods. Based on the physical modelling approach, we categorise them into the following two types:

\begin{figure*}[th!]
    \centering
    \includegraphics[width=\linewidth]{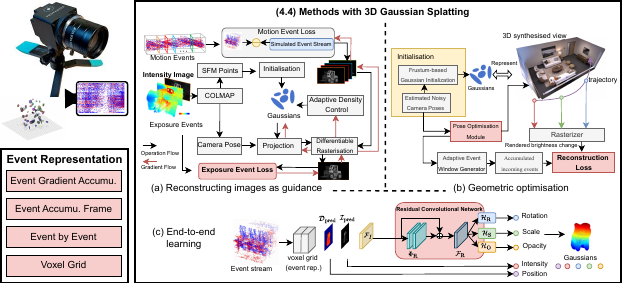}
    \caption{Overview of methods with monocular event cameras that produce 3D Gaussian Splatting. (a) Processing Structure from Yin et al., including Temporal-to-Intensity Mapping to convert exposure events into intensity images, which yield camera trajectories and a sparse point cloud for 3DGS training. (b) Processing Structure from Zahid et al., including physical optimisation by aligning the rendered brightness changes with event-based measurements over adaptive temporal windows. (c) End-to-end learning structure from Wang et al.}
    \label{fig:monocular3}
\end{figure*}

\subsubsection{Brightness difference modelling}
Some relatively early methods supervise NeRF training only by matching the brightness difference between discrete time points, typically over fixed intervals or adjacent event pairs. These methods do not explicitly model the brightness change rate. In 2023, Hwang et al. \cite{[N1]} proposed Ev-NeRF, which aggregates events through multi-view consistency. By estimating the volumetric density and radiance, the method achieves high-quality depth reconstruction and novel view synthesis. In the same year, Rudnev et al. \cite{[N2]} proposed EventNeRF, which employs event-based volumetric rendering in a self-supervised manner to reconstruct high-quality 3D structures and synthesise new views. Later that year, Klenk et al. \cite{[N3]} proposed E-NeRF, which utilises an event-triggered brightness model along with a no-event loss to enable dense reconstruction. In 2024, Bhattacharya et al. \cite{[N4]} proposed EvDNeRF, the first dynamic event-based NeRF. EvDNeRF uses an event-triggered brightness model and a varied batching strategy to achieve high-fidelity dynamic reconstruction. In 2025, Wang et al. \cite{[XN4]} proposed SaENeRF, which normalizes radiance variations based on accumulated event polarities and introduces zero-event regularization losses, enabling artifact-suppressed and photorealistic 3D reconstruction directly from event streams.

\subsubsection{Brightness change rate modelling}
Some later methods incorporate loss functions that explicitly model the brightness change rate, defined as the brightness difference divided by the corresponding time interval. This enables finer temporal resolution, improves robustness to non-uniform motion, and better aligns with the physical characteristics of event triggering. In 2023, Low et al. \cite{[N13]} proposed Robust e-NeRF. This method introduces a more realistic event generation model and two normalised loss functions: one based on contrast-normalised difference and another on target-normalised temporal gradient. As a result, the model no longer requires known contrast thresholds or explicit event accumulation strategies, enabling robust self-supervised learning. In 2024, Wang et al. \cite{[N7]} proposed Physical Priors Augmented EventNeRF, which incorporates motion and geometric priors and adopts a density-guided patch sampling strategy to enhance structural representation. Also in 2024, Low et al. \cite{[XN2]} proposed Deblur e-NeRF, which models pixel bandwidth to account for event motion blur and introduces a threshold-normalised total variation loss, enabling robust 3D reconstruction directly from motion-blurred event streams. In 2025, Feng et al. \cite{[N8]} proposed AE-NeRF, which integrates pose correction and hierarchical architecture to reconstruct NeRFs from sparse and asynchronous event streams accurately. To enhance visual quality, they use a colour correction network to recover RGB images from log-radiance.

\subsection{Methods with 3D Gaussian Splatting}\label{4.4}
\begin{table*}[th!]
    \centering
    \caption{\textbf{Monocular} event camera: \textbf{3DGS}-based Methods}
    \resizebox{\textwidth}{!}{%
    \rowcolors{2}{red!3}{red!10}
    \begin{tabular}{ccccccc}
        \toprule
        \textbf{Author} & \textbf{Model} & \textbf{Yr-Mo} & \textbf{Inputs} & \textbf{Event Rep.} & \textbf{Colourful} & \textbf{Dataset} \\
        \hline
        Wang et al. \cite{[G9]} & EvGGS & 2024-07 & Event stream & Voxel Grid, EAF & \ding{55} & Ev3DS \cite{[G9]}\\        
        Wu et al. \cite{[G6]} & Ev-GS & 2024-09 & Event stream & EGA & \ding{55} & Self-collected \cite{[G6]} \\        
              
        Jeong et al. \cite{[XG2]} & EOGS & 2024-12 & Event stream  & EAF & \ding{55} & EDS \cite{EDS}\\
        Han et al. \cite{[G7]} & Event-3DGS & 2024-12 & Event stream  & Event-by-event & \ding{55} & DeepVoxels \cite{[DeepVoxels]}\\
        Zahid et al. \cite{[G11]} & E-3DGS & 2025-03 & RGB bayer event stream & EAF & \ding{51} & E-3DGS dataset \cite{[G11]}\\
        Yin et al. \cite{[G12]} & E-3DGS & 2025-05 & Event stream & Event-by-event & \ding{55} & EME-3D \cite{[G12]} \\ 
        Zhang et al. \cite{[G10]} & Elite-EvGS & 2025-05 & Event stream & EAF & \ding{55} & Rudnev et al. \cite{[N2]} \\
         Yura et al. \cite{[G5]} & EventSplat & 2025-06 & RGB bayer event stream & EAF & \ding{51} & EDS \cite{EDS}, TUM-VIE \cite{TUM-VIE} \\
        Huang et al. \cite{[G8]} & IncEventGS & 2025-06 & Event stream & EAF & \ding{55} & TUM-VIE \cite{TUM-VIE} \\
        \bottomrule
    \end{tabular}
    }
    \label{tab:mono_gaussian}
\end{table*}

Gaussian Splatting-based methods represent scenes using a set of 3D Gaussian primitives with learnable properties. Each primitive has learnable attributes, including position, covariance, and colour. Similar to NeRF-based methods, most Gaussian Splatting-based methods also use the Event Accumulation Frame as the main input representation of event streams. Table \ref{tab:mono_gaussian} and Figure \ref{fig:monocular3} provide an overview of these methods. Based on the source of supervision and structural guidance, we categorise these methods into the following three types:

\subsubsection{Reconstructing images as guidance}
Some methods use image frames reconstructed from event streams as a source of visual guidance or as an initial structural prior to facilitate the learning process. In 2025, Yin et al. \cite{[G12]} proposed E-3DGS, which uses a novel temporal-to-intensity mapping as visual guidance to facilitate 3D representation learning. The method also incorporates an event-type-specific supervision strategy and a hybrid optimisation approach. Later, Zhang et al. \cite{[G10]} proposed Elite-EvGS, which distils prior knowledge from off-the-shelf event-to-video (E2V) models \cite{[G10-21],[G10-4]}. It uses E2V-generated frames to initialise a coarse 3DGS model and then progressively incorporates raw events to refine scene details through event supervision. Later that year, Yura et al. \cite{[G5]} proposed EventSplat, which combines E2V guided SfM \cite{[MD2-17]} initialisation and spline interpolation. It recovers continuous camera trajectories and achieves high-quality 3D reconstruction.

\subsubsection{Geometric optimisation}
Some methods focus on explicit pose refinement and joint recovery of geometric structure. In 2025, Zahid et al. \cite{[G11]} proposed E-3DGS, which uses frustum-based initialisation to generate an initial Gaussian point cloud. It extracts multi-scale structure and detail using adaptive event windows, and refines camera poses through an event loss to improve trajectory accuracy. In the same year, Huang et al. \cite{[G8]} proposed IncEventGS, which follows a SLAM-inspired framework \cite{[G8-20]} and jointly estimates camera trajectory and 3D structure from event streams.

\noindent\textbf{End-to-end learning: }Some methods focus on directly learning the final outputs from end-to-end learning structure. In 2024, Wang et al. \cite{[G9]} proposed EvGGS that connects depth estimation, intensity reconstruction, and 3D Gaussian parameter regression in a collaborative learning framework. The joint training improves reconstruction accuracy and rendering efficiency. In the same year, Wu et al. \cite{[G6]} proposed Ev-GS that combines neuromorphic imaging with 3DGS, modelling logarithmic brightness changes and enabling fast convergence using only event-based supervision. Later, Han et al. \cite{[G7]} proposed Event-3DGS, which introduces a high-pass filter-based photovoltage estimation module to reduce noise and enhance robustness effectively.

\section{Multimodal Methods with Event Cameras}  \label{5}

Multimodal 3D reconstruction refers to methods that combine event camera data with other sensing modalities \cite{[N9], [XDMT3], [XDMT1]}. These methods aim to enhance reconstruction performance by leveraging the complementary strengths of each modality: the high temporal resolution and low latency of event cameras, and the rich spatial or appearance information provided by other sensors. Compared to stereo and monocular setups, multimodal systems are more flexible in input configurations. They often achieve higher reconstruction accuracy and robustness in low-light, high-speed, or motion-blurred environments.

Early research in this direction typically combines event cameras with active sensing devices, such as structured light projectors \cite{[DMT1]}, which enable high-speed depth recovery through time-coded patterns, and RGB-depth (RGB-D) cameras \cite{[XDMT3]}, which provide readily available depth data for event fusion. Many recent works fuse asynchronous events with frame-based RGB to construct coloured 3D models, or integrate them into advanced neural rendering frameworks such as Neural Radiance Fields and 3D Gaussian Splatting \cite{[N9], [G2]}. These approaches enable photorealistic, temporally-aware 3D reconstruction under challenging real-world conditions.

Based on the output representation and modelling strategy, multimodal methods can be broadly classified into three categories: (\ref{5.1}) methods with traditional 3D outputs such as point clouds, (\ref{5.2}) methods with Neural Radiance Fields, and (\ref{5.3}) methods with 3D Gaussian Splatting.

\subsection{Multimodal Methods with Traditional Outputs} \label{5.1}
\begin{figure*}[ht]
    \centering
    \includegraphics[width=\linewidth]{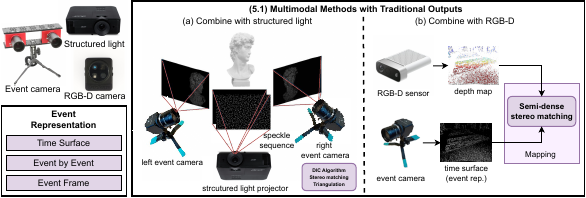}
    \caption{Overview of multimodal methods with traditional outputs. (a) Schematic of event cameras working with structured light projector from Xiao et al. (b) Schematic of a stereo setup of an RGB-D camera and a monocular event camera from Zuo et al.}
    \label{fig:multimodel1}
\end{figure*}
\begin{table*}[t]
    \centering
    \caption{Event cameras in \textbf{multimodal} system: Methods with traditional outputs}
    \resizebox{\textwidth}{!}{%
    \rowcolors{2}{violet!8}{violet!18}
    \begin{tabular}{ccccccc}
        \toprule
        \textbf{Author} & \textbf{Year} & \textbf{Inputs} & \textbf{Priors} & \textbf{Event Rep.} & \textbf{Real-time} & \textbf{Output (Dense?)} \\
        \hline
        Weikersdorfer et al. \cite{[XDMT3]} & 2014 & RGB-D, Event stream & Trajectory & Event-by-event & \ding{51} & Point cloud (\ding{55}) \\
        Matsuda et al. \cite{[XDMT1]} & 2015 & Structured light, Event stream & - & Event-by-event & \ding{51} & 3D depth map (\ding{51}) \\
        Leroux et al. \cite{[DMT1]} & 2018 & Structured light, Event stream & Pose & Time surface & \ding{51} & Point cloud (\ding{51}) \\
        Huang et al. \cite{[DMT2]} & 2021 & Structured light, Event stream & Pose & Event-by-event & \ding{51} & Point cloud (\ding{51}) \\
        Muglikar et al. \cite{[XDMT2]} & 2021 & Structured light, Event stream & - & Event-by-event & \ding{55} & 3D depth map (\ding{51}) \\
        Zuo et al. \cite{[DMT3]} & 2022 & RGB-D, Event stream & Trajectory & Time surface & \ding{51} & Point cloud (\ding{55}) \\
        Xiao et al. \cite{[DMT4]} & 2023 & Structured light, Event stream & Pose & Event frame & \ding{51} & Point cloud (\ding{51}) \\
        Fu et al. \cite{[DMT5]} & 2023 & Structured light, Event stream & Pose & Time surface & \ding{51} & 3D depth map (\ding{55}) \\
        Li et al. \cite{[DMT6]} & 2024 & Structured light, Event stream & Pose & Event-by-event & \ding{51} & 3D depth map (\ding{55}) \\
        \bottomrule
    \end{tabular}
    }
    \label{tab:multimodal_traditional}
\end{table*}
Multimodal methods with traditional outputs recover geometric structures by combining event data with complementary sensors such as structured light or RGB-D cameras. These systems fuse the temporal precision of events with the spatial density of external inputs, enabling robust reconstruction under challenging conditions. Table \ref{tab:multimodal_traditional} and Figure \ref{fig:multimodel1} provide an overview of these methods. As most of them rely on self-collected datasets, datasets are not included in Table~\ref{tab:multimodal_traditional}. Based on the type of complementary modality integrated with event data, we categorise these methods into the following two types:

\begin{figure*}[t]
    \centering
    \includegraphics[width=\linewidth]{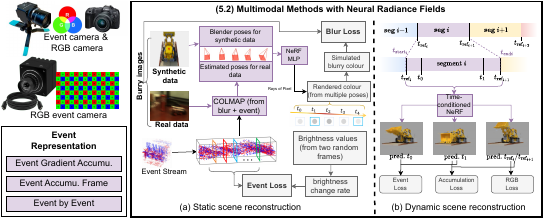}
    \caption{Overview of multimodal methods with Neural Radiance Fields. (a) The method structure from Qi et al. that performs static scene reconstruction. (b) The method structure from Rudnev et al. that performs dynamic scene reconstruction.}
    \label{fig:multimodel2}
\end{figure*}
\begin{table*}[t]
    \centering
    \caption{Event cameras in \textbf{multimodal} system: \textbf{NeRF}-based methods}
    \resizebox{\textwidth}{!}{%
    \rowcolors{2}{violet!8}{violet!18}
    \begin{tabular}{ccccccc}
        \toprule
        \textbf{Author} & \textbf{Model} & \textbf{Yr-Mo} & \textbf{Inputs} & \textbf{Event Rep.} & \textbf{Colourful} & \textbf{Dataset} \\
        \hline
        Qi et al. \cite{[N5]} & E$^2$NeRF & 2023-10 & Blurry RGB, Event stream & EAF & \ding{51} & Self-collected \cite{[N5]} \\
        Ma et al. \cite{[N9]} & Deformable Event-NeRF & 2023-10 & Blurry RGB, Event Stream & Event-by-event & \ding{51} & HS-ERGB \cite{HS-ERGB} \\
        Cannici et al. \cite{[N6]} & Ev-DeblurNeRF & 2024-06 & Blurry RGB, Event stream & EAF & \ding{51} & Ev-DeblurBlender \cite{[N6-D]} \\
        Qi et al. \cite{[N10]} & E$^3$NeRF & 2024-08 & Blurry RGB, Event stream & EGA & \ding{51} & Self-collected \cite{[N10]} \\
        
        Li et al. \cite{[XN5]} & BeNeRF & 2024-10 & Blurry RGB, Event stream & EAF & \ding{51} &  Qi et al. \cite{[N5]}\\
        Qi et al. \cite{[N11]} & EBAD-NeRF & 2024-10 & Blurry RGB, Event stream & EAF & \ding{51} & Self-collected \cite{[N11]} \\
        Chen et al. \cite{[N12]} & Event-ID & 2024-10 & Blurry RGB, Event stream & EAF & \ding{51} & Self-collected \cite{[N12]} \\   
        Tang et al. \cite{[N15]} & LSE-NeRF & 2025-03 & Blurry RGB, Event stream & EAF & \ding{51} & Self-collected \cite{[N15]}, EVIMOv2 \cite{EVIMO2} \\
        Chen et al. \cite{[XN3]} & EvHDR-NeRF & 2025-04 & LDR RGB, Event stream & EAF & \ding{51} & HDR-NeRF dataset \cite{[XN3-D]} \\
        Rudnev et al. \cite{[N14]} & Dynamic EventNeRF & 2025-06 & Blurry RGB, Event stream & EAF & \ding{51} & Self-collected \cite{[N14]} \\
        \bottomrule
    \end{tabular}
    }
    \label{tab:multi_nerf}
\end{table*}

\subsubsection{Combine with Structured light}
Structured light is an active 3D sensing technique that projects coded patterns onto a surface and reconstructs depth via triangulation \cite{StructureLightIntro}. When integrated with event cameras, structured light systems can achieve robust, high-speed depth sensing under challenging conditions. Recent works have introduced several types of encoding and fusion strategies to use both spatial and temporal features of event streams. 

In 2015, Matsuda et al. \cite{[XDMT1]} proposed MC3D, one of the earliest event-based structured light systems. The method correlates laser scan timing with event timestamps, achieving high-precision depth reconstruction with per-pixel single-shot efficiency.
In 2018, Leroux et al. \cite{[DMT1]} proposed event-based structured light systems by projecting frequency-tagged light patterns onto the scene. Each spatial region is illuminated with a unique modulation frequency, and the event camera decodes depth by associating detected frequencies with pixel locations. In 2021, Huang et al. \cite{[DMT2]} combined structured light projection with digital image correlation (DIC) \cite{[DMT2-23],[DMT2-24]} for high-speed scanning. Also in 2021, Muglikar et al. \cite{[XDMT2]} proposed ESL, which estimates depth by maximising spatio-temporal consistency between a laser projector and an event camera. By processing events in local space-time regions, their method improves robustness to noise. ESL accurately estimates depth in challenging scenes but is not real-time due to high computational cost. In 2023, Xiao et al. \cite{[DMT4]} employed alternating binary speckle patterns and DIC-based stereo matching for fast and accurate reconstruction. Also in 2023, Fu et al. \cite{[DMT5]} introduced spatio-temporal coding (STC) with an enhanced matching scheme for improved stereo robustness. In 2024, Li et al. \cite{[DMT6]} proposed eFPSL, using time-frequency analysis to extract high-SNR fringe maps from events and an event-count-based shadow mask to reduce errors.

\subsubsection{Combine with RGB-D}
Some methods fuse event data with RGB-D sensors for improved scene understanding from depth and RGB information. In 2014, Weikersdorfer et al. \cite{[XDMT3]} proposed EB-SLAM-3D, a novel event-based 3D SLAM algorithm using a D-eDVS, enabling low-power, low-latency mapping with a sparse voxel grid at 20× real-time speed. In 2022, Zuo et al. \cite{[DMT3]} proposed DEVO, combining time surface maps from events and depth supervision from a calibrated sensor. Their system performs semi-dense 3D-2D edge alignment to estimate poses and incrementally build point clouds under fast motion and poor lighting.

\subsection{Multimodal Methods with Neural Radiance Fields} \label{5.2}

\begin{figure*}[t]
    \centering
    \includegraphics[width=\linewidth]{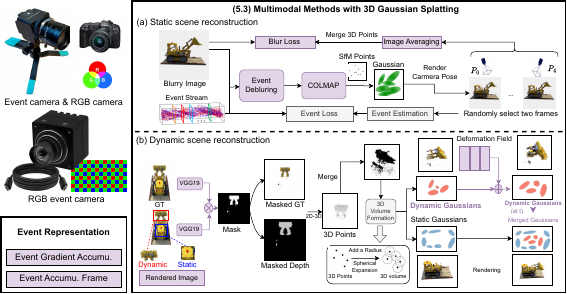}
    \caption{Overview of multimodal methods with 3D Gaussian Splatting. (a) The method structure from Deguchi et al. that performs static scene reconstruction. (b) The method structure from Xu et al. that performs dynamic scene reconstruction.}
    \label{fig:multimodel3}
\end{figure*}
\begin{table*}[t]
    \centering
    \caption{Event cameras in \textbf{multimodal} system: \textbf{3DGS}-based methods}
    \resizebox{\textwidth}{!}{%
    \rowcolors{2}{violet!8}{violet!18}
    \begin{tabular}{ccccccc}
        \toprule
        \textbf{Author} & \textbf{Model} & \textbf{Yr-Mo} & \textbf{Inputs} & \textbf{Event Rep.} & \textbf{Colourful} & \textbf{Dataset} \\
        \hline
        Yu et al. \cite{[G2]} & EvaGaussians & 2024-05 & Blurry RGB, Event stream & EGA & \ding{51} & Self-collected \cite{[G2]} \\
        Xiong et al. \cite{[G4]} & Event3DGS & 2024-06 & Blurry RGB, Event stream & EGA & \ding{51} & Self-collected \cite{[G4]} \\
        Weng et al. \cite{[G3]} & EaDeblur-GS & 2024-07 & Blurry RGB, Event stream & EGA & \ding{51} & Qi et al. \cite{[N5]} \\
        Liao et al. \cite{[G13]} & EF-3DGS & 2024-10 & Blurry RGB, Event stream & EAF & \ding{51} & Tanks and Temples \cite{Tanks} \\
        Deguchi et al. \cite{[G1]} & E2GS & 2024-10 & Blurry RGB, Event stream & EAF & \ding{51} & Self-collected \cite{[G1]} \\
        Xu et al. \cite{[G15]} & EventBoosted-3DGS & 2024-11 & Blurry RGB, Event stream & EGA & \ding{51} & Self-collected \cite{[G15]} \\
        Huang et al. \cite{[XG1]} & Ev3DGS & 2024-12 & Blurry RGB, Event stream & EAF & \ding{51} & Qi et al. \cite{[N5]}\\
        Wu et al. \cite{[G14]} & SweepEvGS & 2024-12 & Static RGB, Event stream & EGA & \ding{55} & Self-collected \cite{[G14]} \\
        Lee et al. \cite{[XG3]} & Sensor Fusion Splatting & 2025-02 & RGB, Event stream, Depths & EGA & \ding{51} & Self-collected \cite{[XG3]} \\
        Matta et al. \cite{[XG4]} & BeSplat & 2025-03 & Blurry RGB, Event stream & EAF & \ding{51} & BeNeRF dataset \cite{[XG4-D]}\\
        Deng et al. \cite{[XG5]} & EBAD-Gaussian & 2025-04 & Blurry RGB, Event stream & EAF & \ding{51} & Qi et al. \cite{[N11]} \\
        Lee et al. \cite{[XG6]} & DiET-GS & 2025-06 & Blurry RGB, Event stream & EGA & \ding{51} & Cannici et al. \cite{[N6]} \\
        \bottomrule
    \end{tabular}
    }
    \label{tab:multi_gaussian}
\end{table*}

Multimodal Event-based NeRF methods combine event streams with RGB images to achieve high-quality coloured 3D reconstruction, leveraging the temporal precision of events and the rich appearance information of RGB frames to enhance structural accuracy and visual fidelity under motion blur and low-light conditions. Table \ref{tab:multi_nerf} and Figure \ref{fig:multimodel2} provide an overview of these methods. Based on the type of scene being reconstructed, we categorise these methods into the following two types:

\subsubsection{Static scene reconstruction}
Some methods focus on static scene reconstruction by integrating event supervision with blurry RGB images \cite{[N5],[N6],[N10],[N15]}, or by designing physically inspired mechanisms \cite{[N11],[N12]} to improve NeRF training stability and output sharpness. In 2023, Qi et al. \cite{[N5]} proposed E²NeRF, which introduces a blur rendering loss and an event rendering loss to improve NeRF training under motion blur and low-light settings. In 2024, Cannici et al. \cite{[N6]} proposed Ev-DeblurNeRF, which uses the Event Double Integral (EDI) \cite{[N6-29]} and a learnable event camera response function (eCRF) to reconstruct sharp NeRFs. It performs well under severe motion blur. Later that year, Qi et al. \cite{[N10]} proposed E3NeRF, which combines blurry images and events with spatial-temporal attention \cite{guo2022attention}. It introduces an event-enhanced rendering loss to guide learning and improves robustness in non-uniform motion and low-light scenes. Later, Li et al. \cite{[XN5]} proposed BeNeRF, which jointly reconstructs photorealistic 3D scenes and estimates camera motion from a single blurry RGB image and event stream, using an event rendering loss and B-spline trajectory representation for end-to-end optimisation. Around the same time, Qi et al. \cite{[N11]} proposed EBAD-NeRF, which introduces event-driven bundle adjustment. It jointly optimises NeRF and camera poses using an intensity-change event loss and a photometric blur loss. Also in 2024, Chen et al. \cite{[N12]} proposed Event-ID, which is the first event-driven framework for intrinsic decomposition. It combines an event-based reflectance model and a multi-view strategy to recover geometry, materials, and lighting under extreme conditions. In 2025, Tang et al. \cite{[N15]} proposed LSE-NeRF, which models sensor response differences with per-time embeddings and event-based reflectance mapping to recover high-quality NeRFs without strict alignment. Later, Chen et al. \cite{[XN3]} proposed EvHDR-NeRF, which models a radiance-based relationship that accounts for exposure time and the camera response function (CRF), enabling HDR 3D reconstruction from single-exposure LDR images and event streams.

\subsubsection{Dynamic scene reconstruction}
Some methods reconstruct dynamic scenes by combining event streams with deformable or time-conditioned NeRF frameworks. In 2024, Ma et al. \cite{[N9]} proposed DE-NeRF, which is the first deformable NeRF framework that fuses RGB images and event data. It combines continuous pose estimation and a learnable deformation field. The method enables high-quality dynamic scene reconstruction and novel view synthesis. In 2025, Rudnev et al. \cite{[N14]} proposed Dynamic EventNeRF, which uses time-conditioned NeRF models and introduces an event accumulation damping mechanism. The method achieves photorealistic synthesis in low-light and high-speed dynamic scenes using only event cameras and sparse blurry RGB frames.

\subsection{Multimodal Methods with 3D Gaussian Splatting}  \label{5.3}

\begin{figure*}[t]
    \centering
    \includegraphics[width=0.8\linewidth]{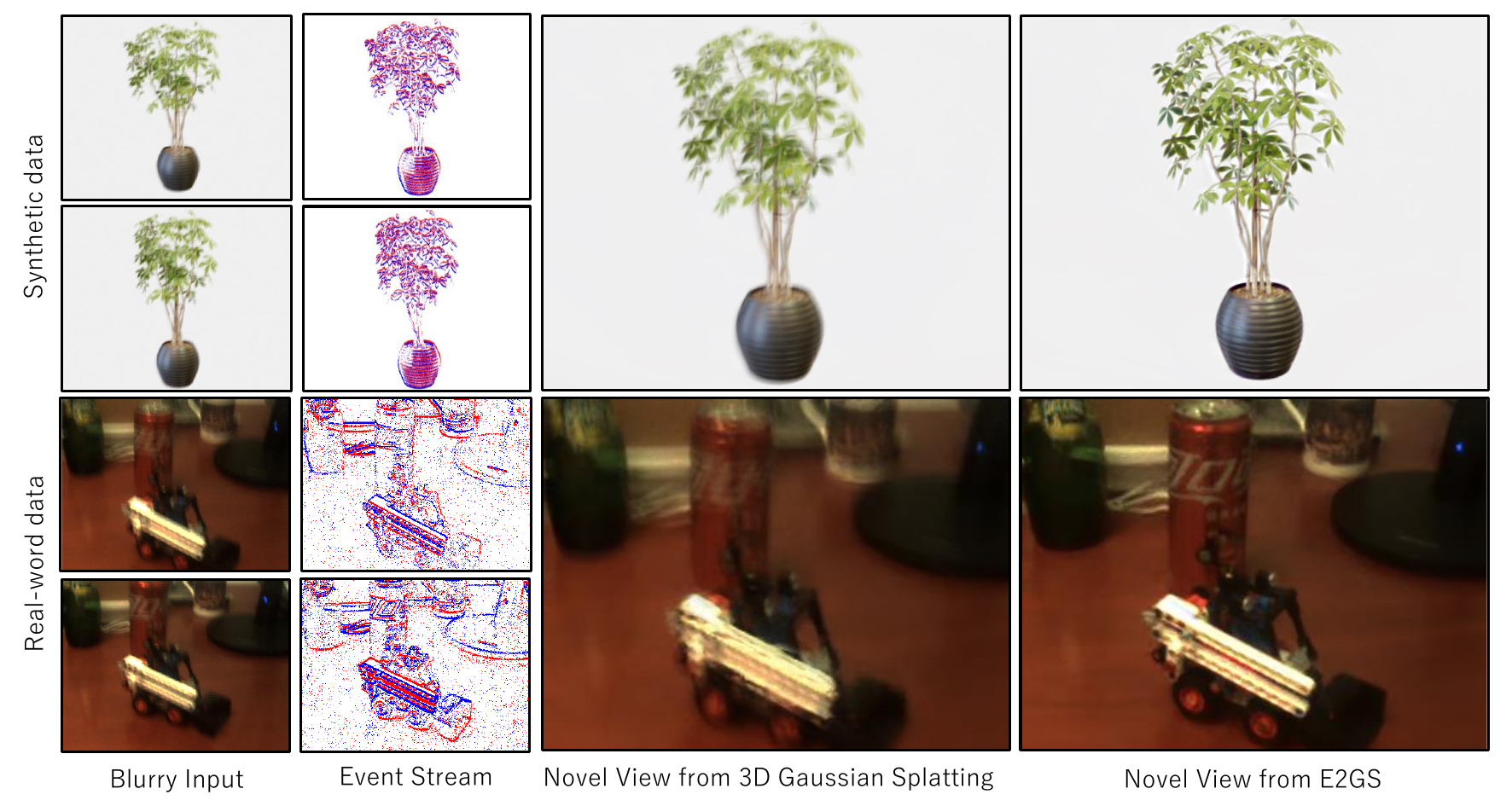}
    \caption{Multimodal 3D Gaussian reconstruction based on blurry RGB and event data originates from the E2GS method under CC BY 4.0. @IEEE 2024. The event modality assists in deblurring scene rendering.}
    \label{fig:G1}
\end{figure*}

Multimodal Event-based Gaussian Splatting methods achieve high-quality 3D reconstruction by combining event streams with RGB images. These multimodal approaches leverage the high temporal resolution of events and the rich appearance information of RGB frames to improve geometric accuracy and colour fidelity. Table \ref{tab:multi_gaussian} and Figure \ref{fig:multimodel3} provide an overview of these methods. Based on the type of scene being reconstructed, we categorise these methods into the following two types:

\subsubsection{Static scene reconstruction}
Some methods focus on static scene reconstruction by using event streams to compensate for motion blur and improve 3D reconstruction from blurry images. In 2024, Yu et al. \cite{[G2]} proposed EvaGaussians, which introduces learnable camera pose offsets and jointly optimises blurry image trajectories and 3D Gaussians. It improves reconstruction accuracy under severe motion blur conditions. Later that year, Weng et al. \cite{[G3]} proposed EaDeblur-GS, incorporating an Adaptive Deviation Estimator (ADE) \cite{[G3-9],[G3-2]} network and blur-specific losses. It enables real-time and high-fidelity 3D reconstruction from extremely blurred inputs. In the same year, Deguchi et al. \cite{[G1]} proposed E2GS, which combines Event-based Double Integral (EDI) \cite{[G1-19]} and event rendering loss to enhance blurry image recovery and achieve 3D Gaussian reconstruction. Also in 2024, Huang et al. \cite{[XG1]} proposed Ev3DGS, which introduces blur and event rendering losses to guide 3D Gaussian Splatting, enabling fast and accurate colour reconstruction from blurry RGB images and event streams. In 2025, Matta et al. \cite{[XG4]} proposed BeSplat, which reconstructs sharp and colour-consistent 3D scenes from a single blurry RGB image and event stream by jointly optimising scene representation and camera motion with event-guided spatio-temporal lifting. Later, Deng et al. \cite{[XG5]} proposed EBAD-Gaussian, which leverages complementary image and event modalities and introduces event-driven bundle adjustment with motion blur modelling to jointly optimise Gaussians and camera poses, enabling sharp and physically consistent 3D reconstruction under severe motion blur. In the same year, Lee et al. \cite{[XG6]} proposed DIET-GS, which also integrates the EDI prior with a pre-trained diffusion model. It employs a two-stage training strategy to constrain 3D Gaussians with event data for accurate colour recovery, while leveraging the diffusion prior to further refine fine-grained details and enhance edge sharpness. Figure \ref{fig:G1} provides a visualisation of its result compared to RGB-based 3D Gaussian Splatting \cite{[Gaussian]}.

A method uses grayscale images as auxiliary input. Compared to other multimodal approaches that combine RGB images, this event-driven strategy is more lightweight. In 2024, Wu et al. \cite{[G14]} proposed SweepEvGS, which is the first 3DGS framework that reconstructs macro- and micro-scale radiance fields from a single sweep of event data. It utilises event-based supervision and structure loss to achieve efficient and robust novel view synthesis.

\subsubsection{Dynamic scene reconstruction}
Some methods achieve dynamic 3D reconstruction either by modelling continuous time supervision from event streams \cite{[G4], [G13]} or by introducing non-rigid deformation fields to explicitly represent scene dynamics \cite{[G15], [XG3]}. In 2024, Xiong et al. \cite{[G4]} proposed Event3DGS, which introduces sparsity-aware sampling and a progressive training strategy. It reconstructs geometrically consistent and efficient 3D structures from event streams under high-speed egomotion. In the same year, Liao et al. \cite{[G13]} proposed EF-3DGS, which is the first event-aided 3DGS framework that supports free-trajectory rendering in dynamic scenes. It enhances reconstruction accuracy and robustness in dynamic scenes by combining a Linear Event Generation Model (LEGM) \cite{[G13-10],[G13-13],[G13-56]} with a contrast maximisation-based image sharpening strategy \cite{[G13-11],[G13-15],[G13-34],[G13-38]}. Also in 2024, Xu et al. \cite{[G15]} proposed Event-Boosted Deformable 3D Gaussians, which is the first deformable 3DGS framework incorporating event cameras. It applies a GS-threshold joint modelling strategy and a dynamic-static decomposition mechanism to achieve high-quality and efficient dynamic reconstruction. In 2025, Lee et al. \cite{[XG3]} proposed Sensor Fusion Splatting, which fuses RGB images, event streams, and depth maps from an RGB-D camera with deformable Gaussians and modality-specific losses, enabling high-quality 3D reconstruction in dynamic scenes.

\section{Datasets} \label{6}

Although numerous event-based vision datasets have been introduced, only a limited subset is suitable for 3D reconstruction, and many of them are not specially designed for 3D reconstruction. Many recent works rely on private datasets (e.g., \cite{[G3]}) or synthesise events from RGB videos using simulators without releasing the resulting data (e.g., \cite{[G2]}, \cite{[G15]}, \cite{[S5]}). As a result, only a small number of publicly available datasets provide the dense depth, stereo disparity, or sub-millimetre trajectories required for rigorous evaluation of reconstruction performance. Table~\ref{tab:event_datasets} summarises the main publicly available datasets. Table \ref{tab:event_datasets_visualisation} provides example visualisation of sensors, RGB frames, event frames, and labels from the representative datasets. In Table \ref{tab:event_datasets_visualisation}, all the synthetic datasets are created by Blender \cite{blender}, and parts of the figures in the table are referenced from Ghosh et al. with permission \cite{ghosh2024event}). 

These datasets naturally fall into four categories based on their data acquisition setting, type of geometric supervision, and target reconstruction tasks: (1) outdoor navigation datasets with LiDAR or visual-inertial ground truth, (2) indoor object and human reconstruction datasets with motion capture or laser scans, (3) synthetic datasets with dense annotations, and (4) photometric benchmarks with accurate camera poses but without metric depth.

\begin{table*}[th!]
    \centering
    \caption{Publicly available datasets for 3D reconstruction tasks with event cameras. (Abbr.: E = event stream, RGB = image frames, Li = LiDAR, IMU = inertial unit, Vcn = Vicon.)}
    \resizebox{\textwidth}{!}{%
    \rowcolors{2}{gray!3}{gray!10}
    \begin{tabular}{lcccccc}
        \hline
        \textbf{Dataset} & \textbf{Venue (Year)} & \textbf{Type} & \textbf{Sensors / Resolution} & \textbf{Label} & \textbf{Size}  \\
        \hline
                MVSEC \cite{MVSEC}                  & RA-L (2018) & Real      & Stereo 346$\times$260 E, Li, IMU, GPS                     & Point Cloud        & 30 GB        \\
        DHP19 \cite{DHP19}                  & CVPR (2019) & Real      & 4$\times$346$\times$260 E, Vcn                            & 13‑joint skeleton  & 30 GB     \\
        DSEC \cite{[S7-13]}\cite{DSEC2}     & RA-L (2021) & Real      & 2$\times$640$\times$480 E, 2$\times$1.4 MP RGB, Li, IMU  & Depth Map (LiDAR)  & 150 GB     \\
        TUM‑VIE \cite{TUM-VIE}              & IROS (2021) & Real      & Stereo 1280$\times$720 E, IMU 200 Hz, Vcn                & Grayscale Frames   & 300 GB      \\
        ViViD++ \cite{ViViD++}              & RA-L (2022) & Real      & Mono E + Thermal + Li                                     & Pose               & 200 GB       \\
        MOEC‑3D \cite{[MD5]}                & ECCV (2022) & Real      & Mono E, laser mesh                                        & Mesh               & 30 GB        \\
        EVIMO‑2 \cite{EVIMO2}              & Arxiv (2022) & Real      & 3$\times$640$\times$480 E, RGB, 2 IMU, Vcn                & Object Pose        & 350 GB       \\

        EventScape \cite{EventScape}        & RA-L (2021) & Synthetic & CARLA (E+RGB)                                             & Dense Depth Maps   & 70 GB        \\
        EventNeRF \cite{[N3]}               & CVPR (2023) & Synthetic & Mono E + RGB Refs                                        & RGB Frame          & 18 GB        \\
        SynthEVox3D \cite{[Dense]}          & ICVR (2023) & Synthetic & Mono E (E2V) from Blender Simulator                                         & Voxel Grid         & 32 GB       \\
        SEVD \cite{SEVD}                    & Arxiv (2024) & Synthetic & CARLA multiview E+RGB                                     & Depth Map, Masks   & 300 GB  \\

        DAVIS240C \cite{DAVIS240C}          & IJRR (2017) & Real + Synthetic & Mono 240$\times$180 E, IMU                                & Pose               & 10 GB    \\
        \hline
    \end{tabular}}
    \label{tab:event_datasets}
\end{table*}

\subsection{Outdoor navigation datasets}  
This category comprises ego-centric navigation sets, typified by DSEC \cite{[S7-13]}\cite{DSEC2}, MVSEC \cite{MVSEC}, TUM-VIE \cite{TUM-VIE}, and ViViD++ \cite{ViViD++}. These datasets align stereo events with LiDAR or tightly fused visual–inertial trajectories, yielding metre-scale ground truth that is indispensable for depth estimation and visual–inertial SLAM in outdoor environments. For urban driving scenarios, DSEC remains the reference standard, whereas TUM-VIE offers higher-resolution sensors and sub-centimetre inertial poses for precision studies.

\subsection{Indoor object and human reconstruction}  
This group targets object- and human-centric reconstruction in controlled indoor studios. EVIMO-2 \cite{EVIMO2} and MOEC-3D \cite{[MD5]} provide millimetre-accurate meshes or per-pixel depth obtained from multi-camera motion capture or laser scanning, making them well suited to investigations of articulated or rigid-body shape recovery. DHP19 \cite{DHP19} is currently the sole public resource for full-body event capture, delivering four DAVIS346 views and millimetre-scale skeletons for non-rigid human modelling.

\subsection{Synthetic datasets} 
Where real scenes lack dense supervision, synthetic datasets provide idealised ground truth at scale. SynthEVox3D \cite{[Dense]}, SEVD \cite{SEVD}, and EventScape \cite{EventScape} render voxel occupancies, semantic masks, and depth maps that enable large-scale pre-training and controlled ablation studies. While they offer reliable supervision for training, bridging the domain gap to real-world sparse inputs remains an open challenge, often addressed through augmentation strategies or adaptation techniques.

\begin{table*}[hp]
  \centering
  \caption{Representative datasets used for event-based 3D Tasks.}
  \label{tab:event_datasets_visualisation}

  \begin{adjustbox}{max width=\textwidth}
  {\small
   \setlength{\tabcolsep}{2pt}
   \begin{tabular}{
     >{\centering\arraybackslash}m{0.2\textwidth}
     >{\centering\arraybackslash}m{0.22\textwidth}
     >{\centering\arraybackslash}m{0.22\textwidth}
     >{\centering\arraybackslash}m{0.22\textwidth}
     >{\centering\arraybackslash}m{0.22\textwidth}
   }
     \toprule
     \textbf{Dataset} & \textbf{Sensors}
                     & \textbf{Frames}
                     & \textbf{Events}
                     & \textbf{Labels} \\
     \midrule

     \makecell{DSEC\\~\cite{[S7-13]}\cite{DSEC2}} &
     \includegraphics[width=0.95\linewidth]{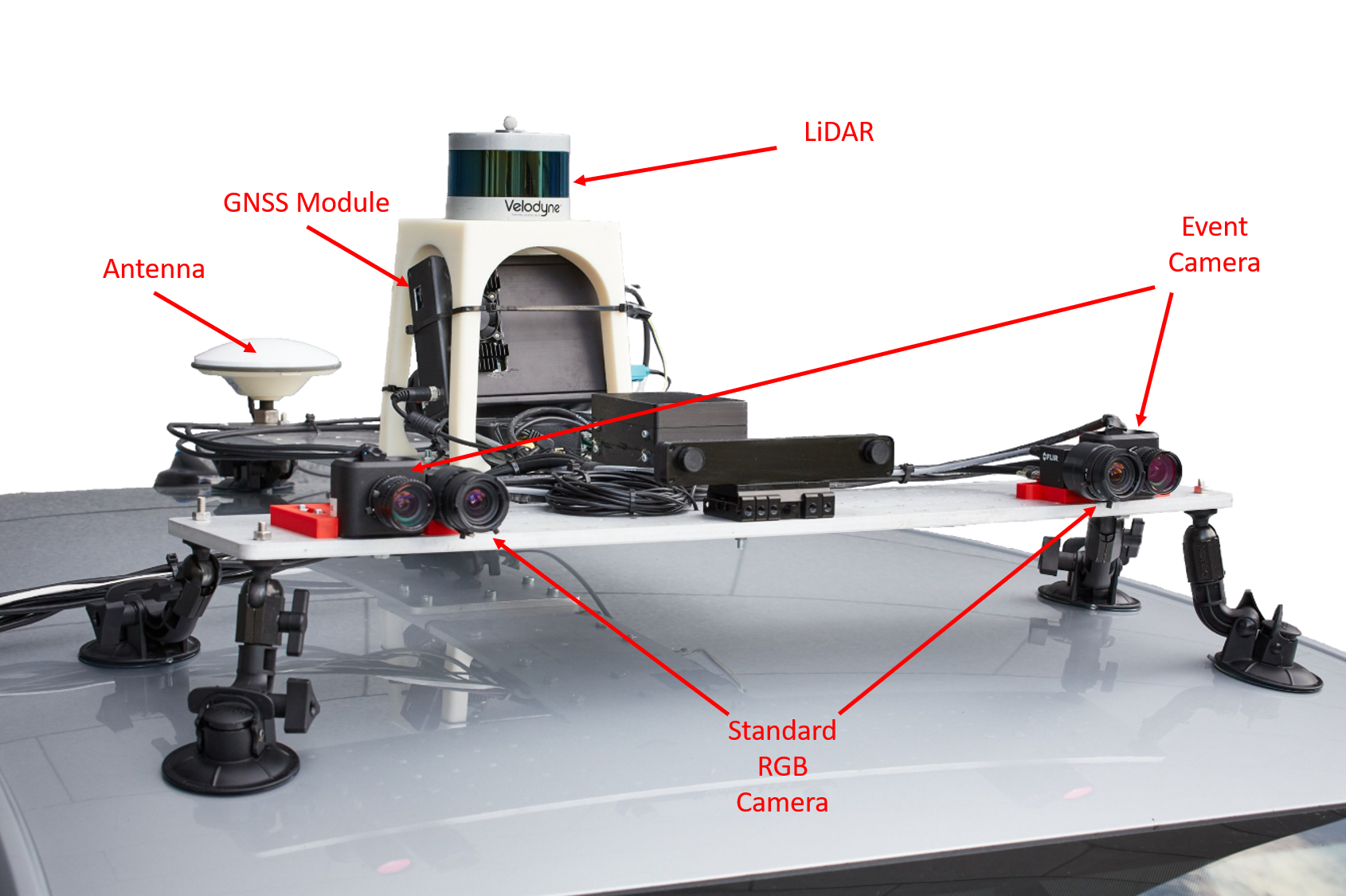} &
     \includegraphics[width=0.95\linewidth]{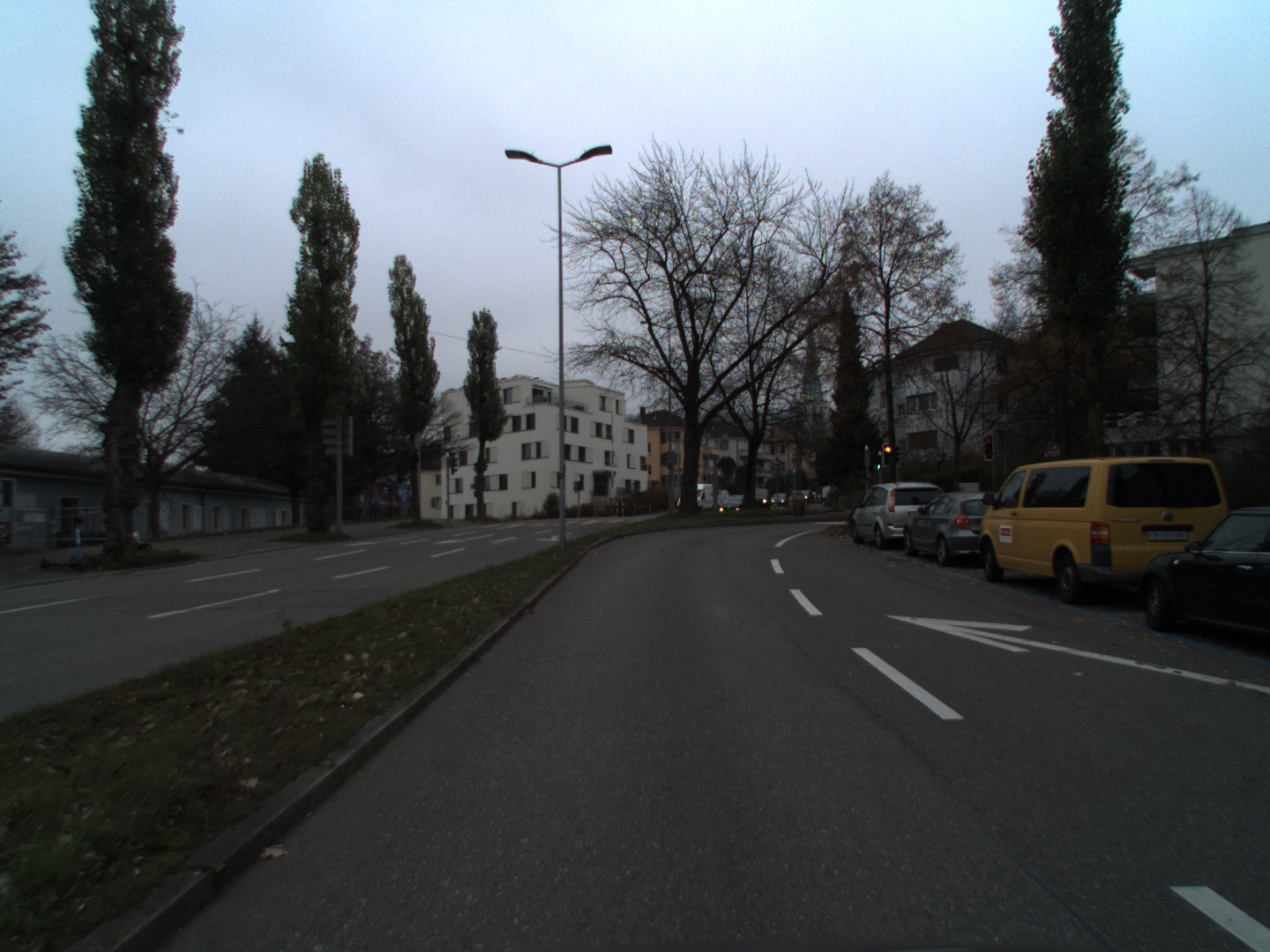} &
     \includegraphics[width=0.95\linewidth]{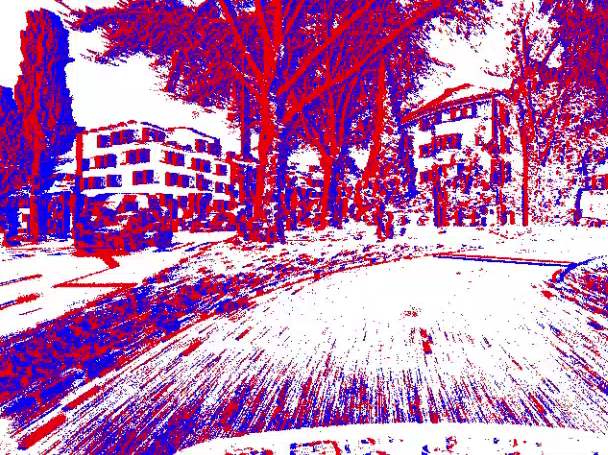} &
     \includegraphics[width=0.95\linewidth]{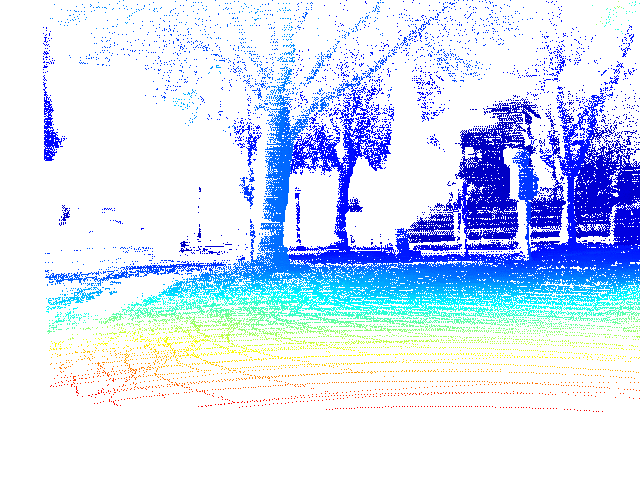} \\

     \makecell{MVSEC\\~\cite{MVSEC}} &
     \includegraphics[width=0.95\linewidth]{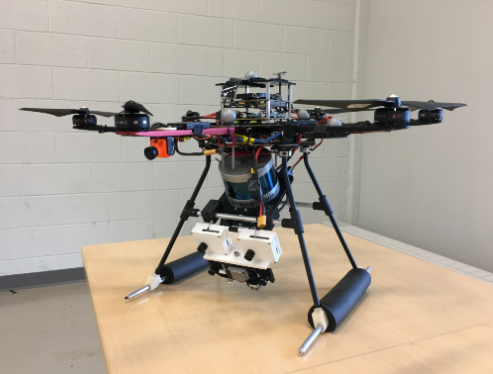} &
     \includegraphics[width=0.95\linewidth]{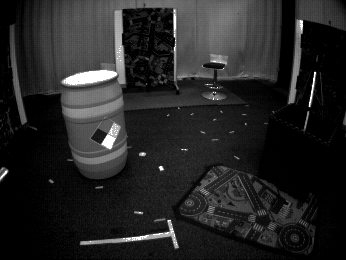} &
     \includegraphics[width=0.95\linewidth]{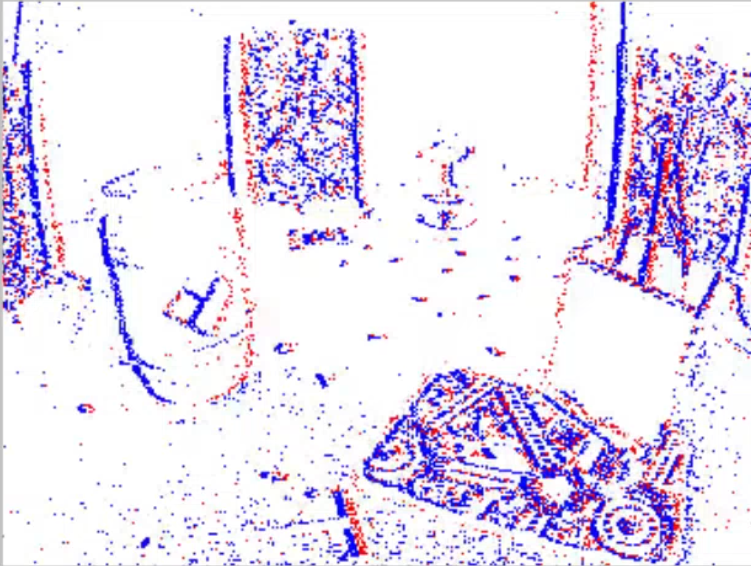} &
     \includegraphics[width=0.95\linewidth]{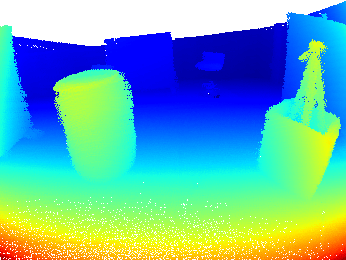} \\

     \makecell{TUM-VIE\\~\cite{TUM-VIE}} &
     \includegraphics[width=0.95\linewidth]{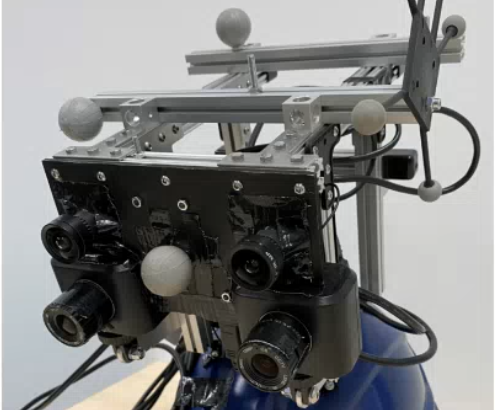} &
     \includegraphics[width=0.95\linewidth]{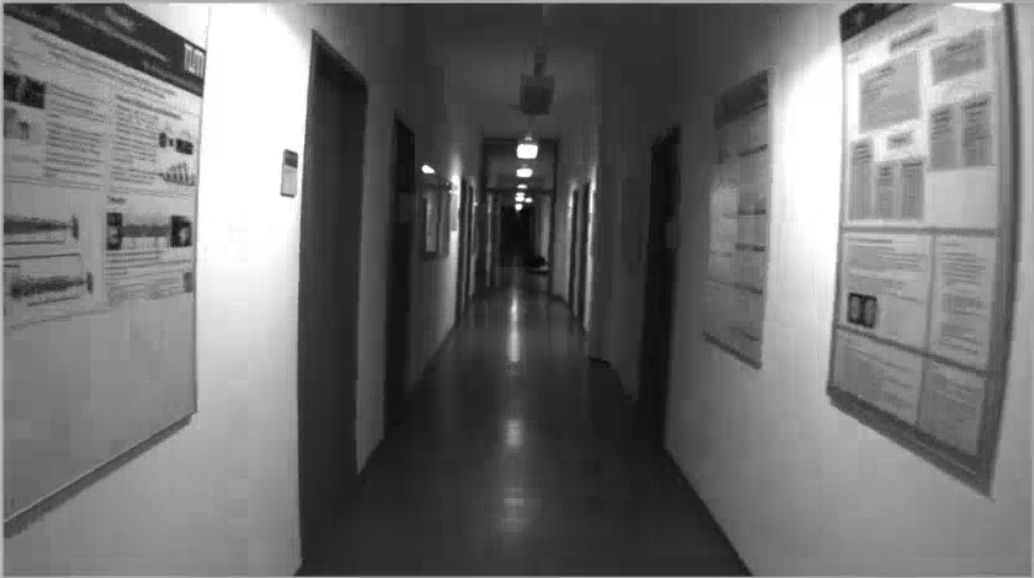} &
     \includegraphics[width=0.95\linewidth]{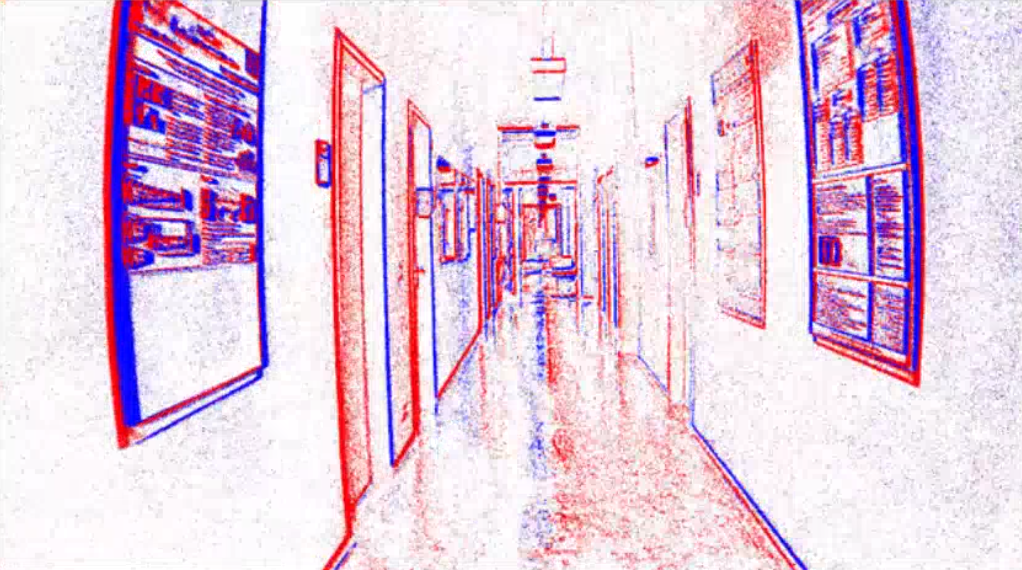} &
     \includegraphics[width=0.95\linewidth]{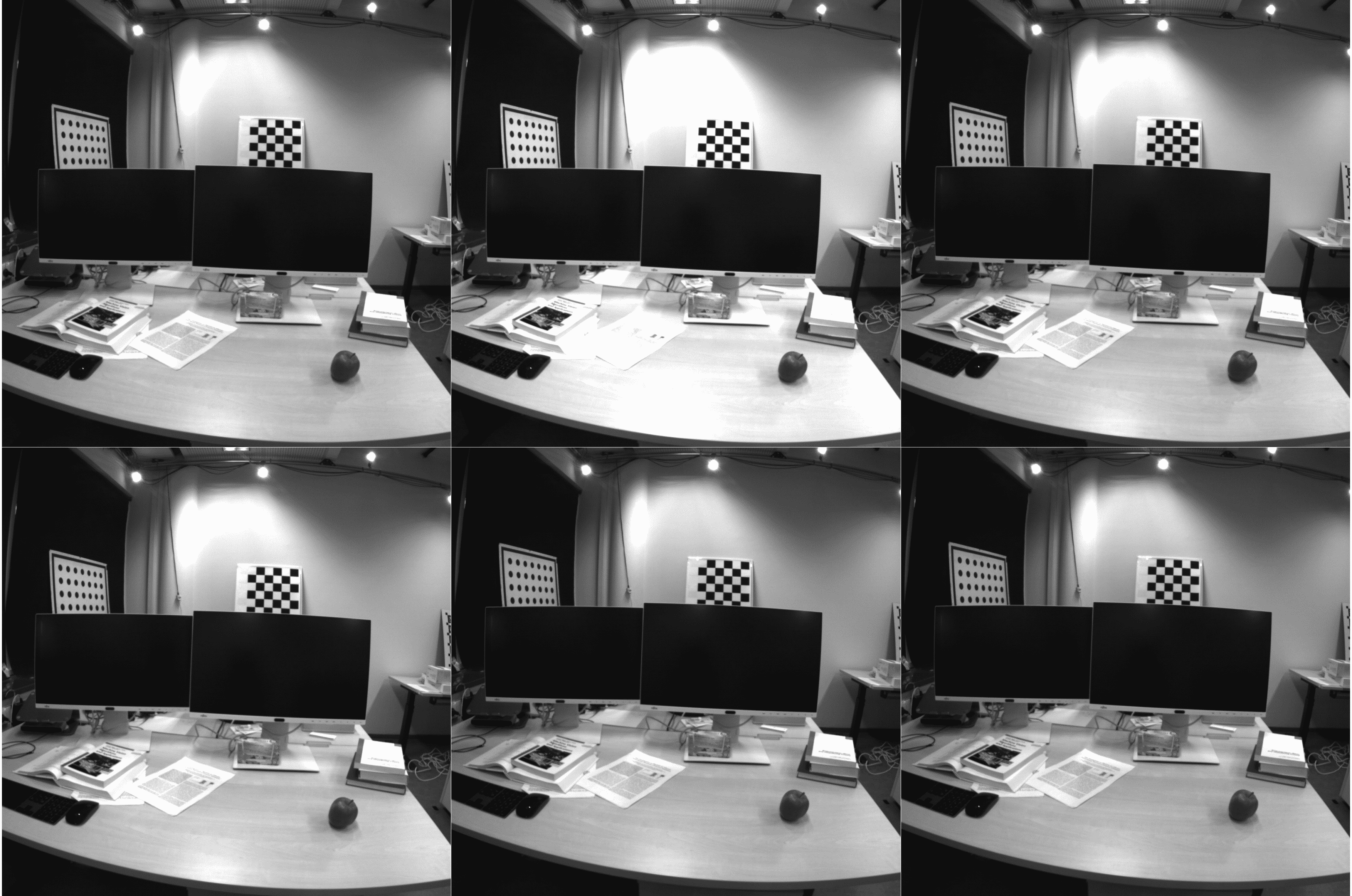} \\
     %&\makecell{IMU\\Data} \\

     \makecell{ViViD++\\~\cite{ViViD++}} &
     \includegraphics[width=0.95\linewidth]{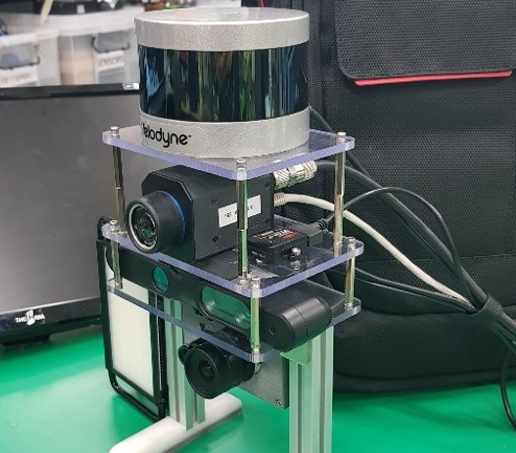} &
     \includegraphics[width=0.95\linewidth]{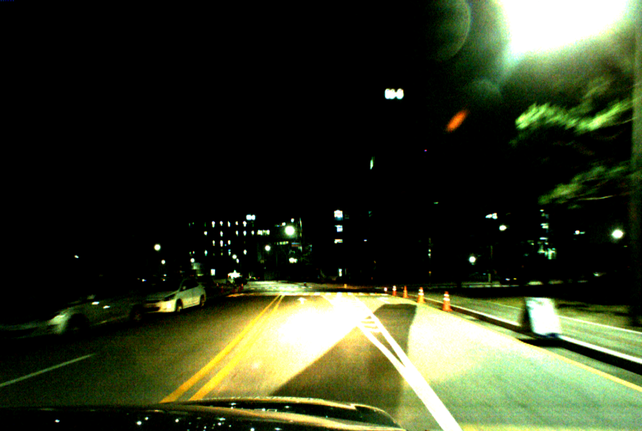} &
     \includegraphics[width=0.95\linewidth]{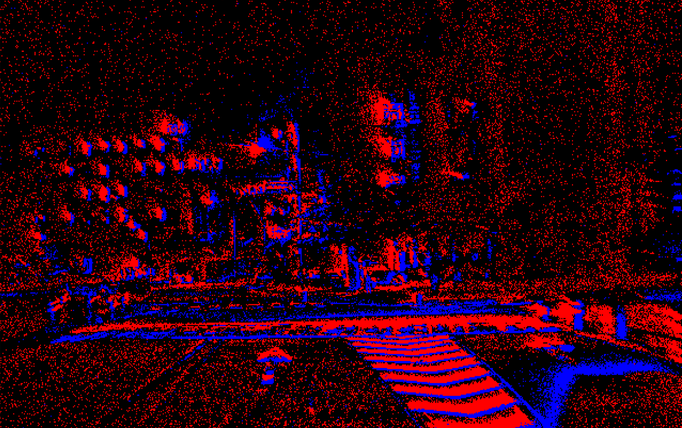} &
     \includegraphics[width=0.95\linewidth]{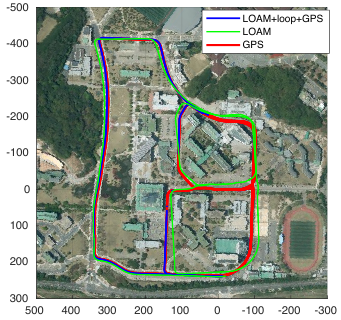} \\

     \makecell{EV-IMO2\\~\cite{EVIMO2}} &
     \includegraphics[width=0.95\linewidth]{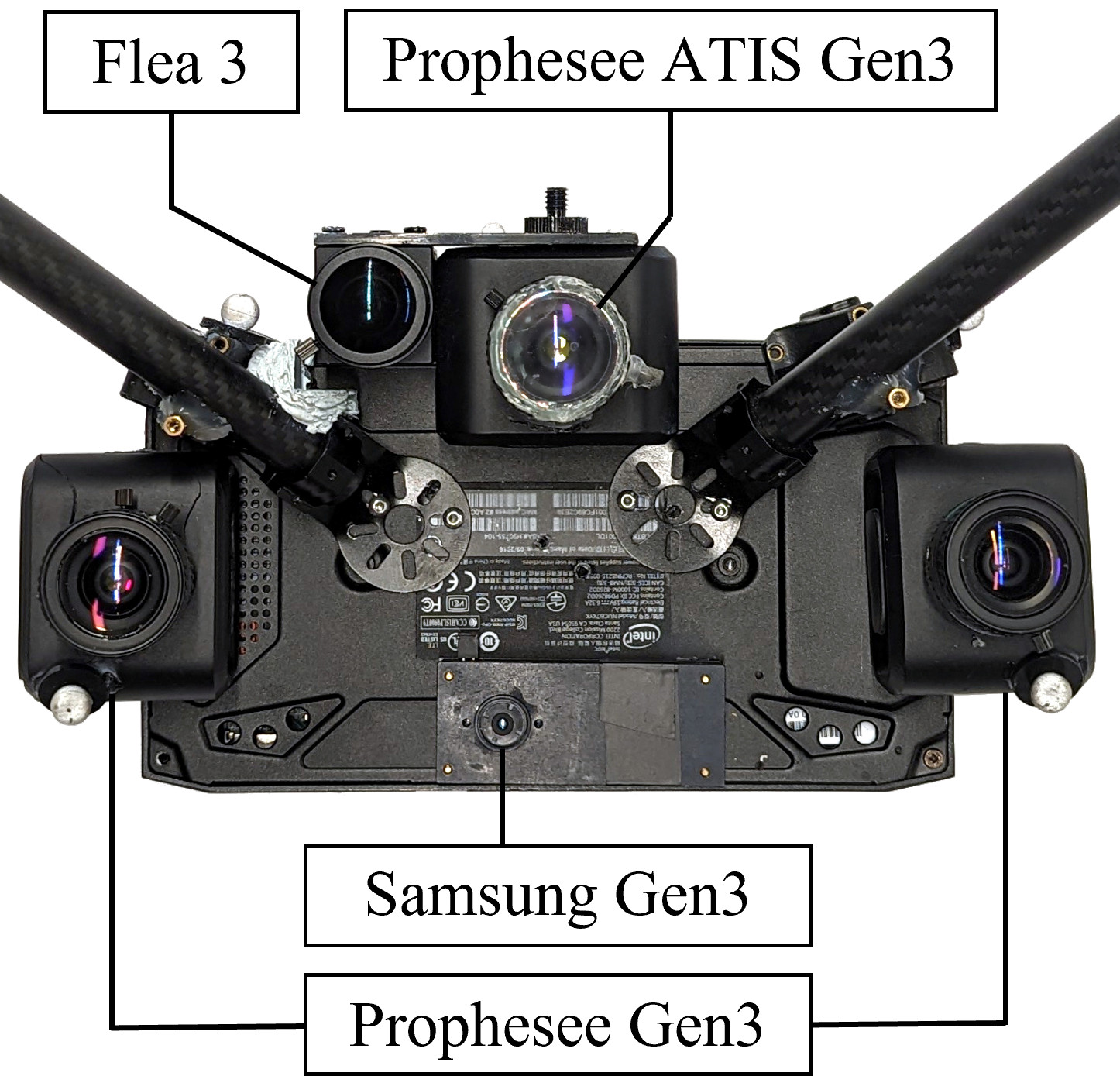} &
     \includegraphics[width=0.95\linewidth]{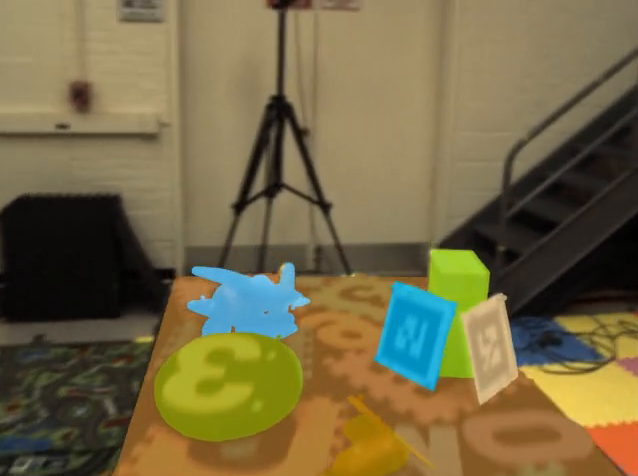} &
     \includegraphics[width=0.95\linewidth]{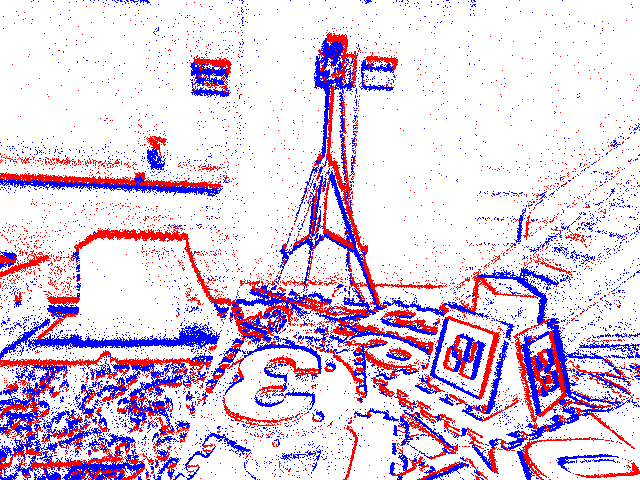} &
     \includegraphics[width=0.95\linewidth]{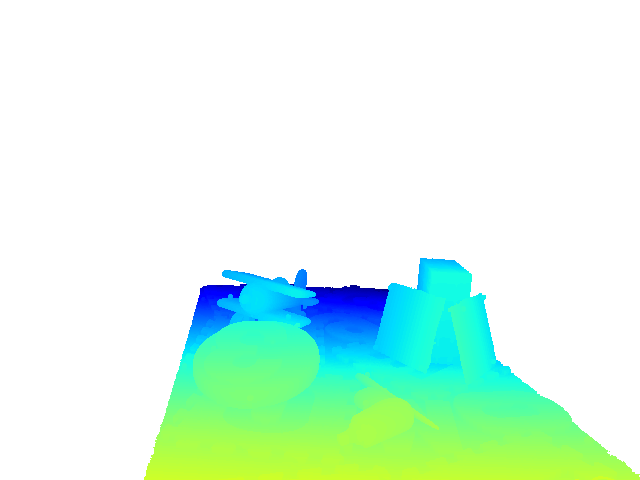} \\

     \makecell{SynthEVox3D\\~\cite{[Dense]}} &
     %\makecell{Synthetic\\Dataset} &
     \includegraphics[width=0.95\linewidth]{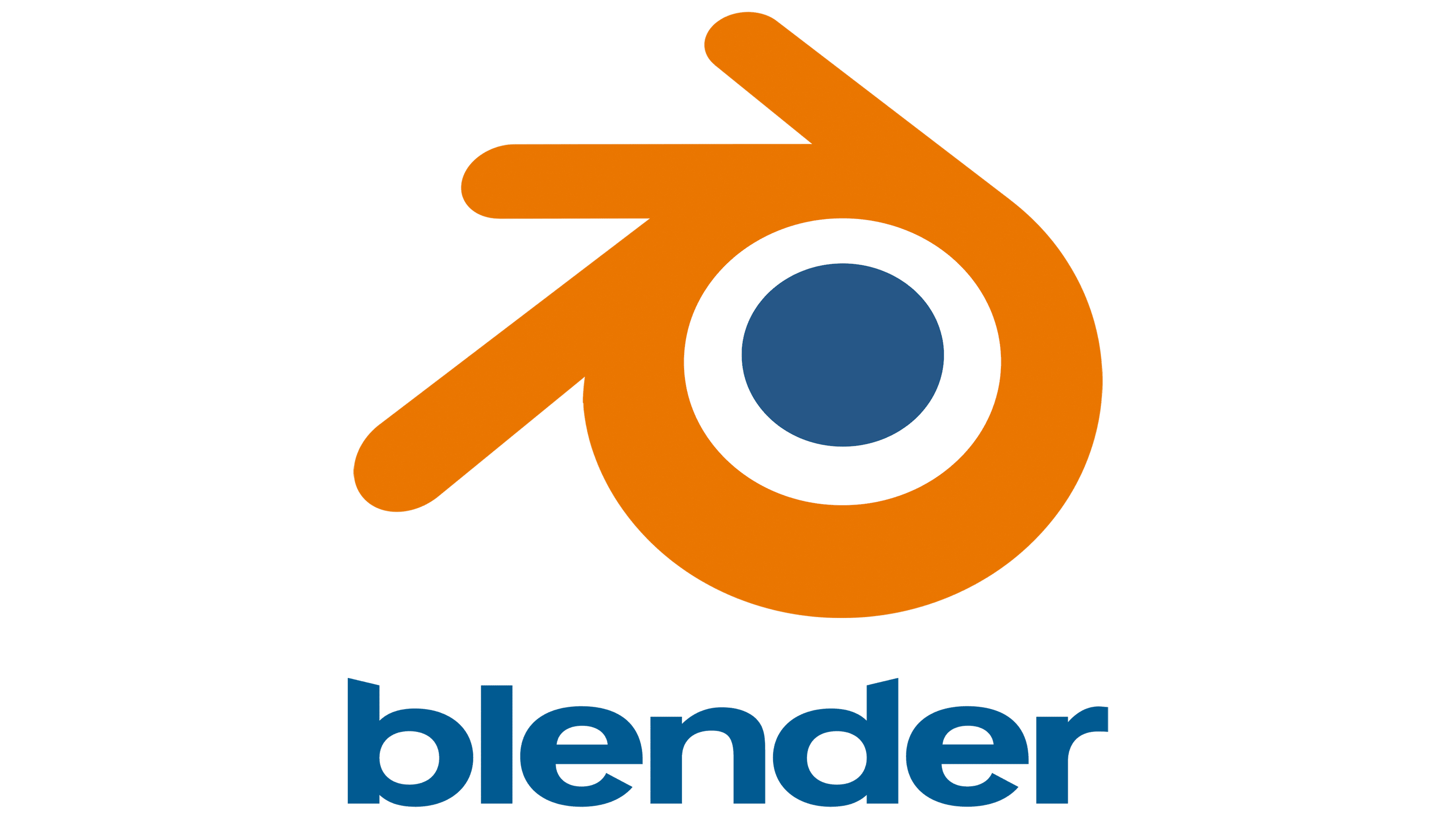} &
     \includegraphics[width=0.95\linewidth]{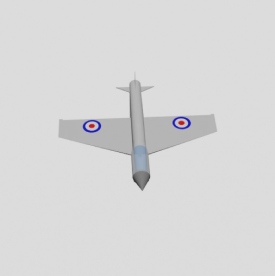} &
     \includegraphics[width=0.95\linewidth]{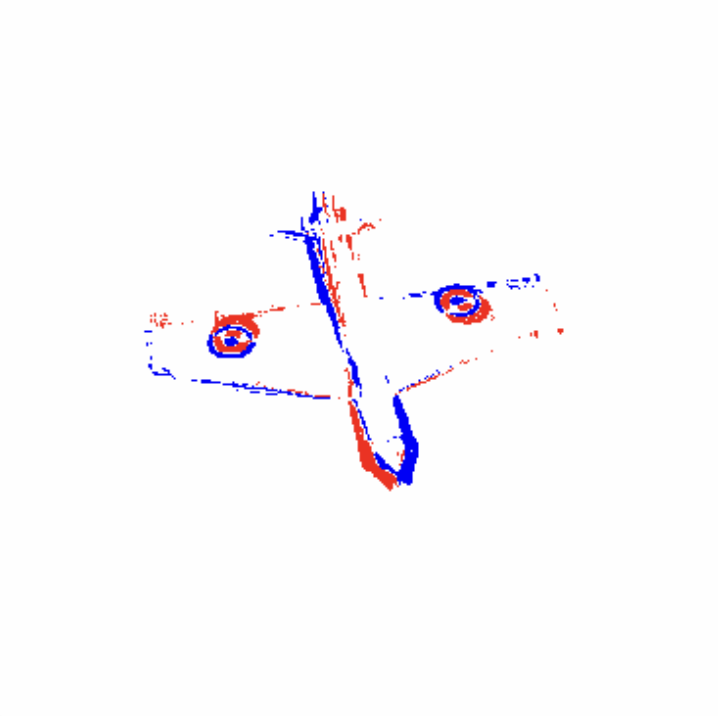} &
     \includegraphics[width=0.95\linewidth]{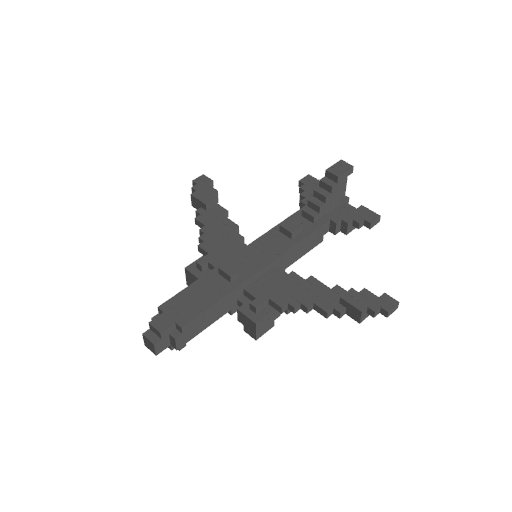} \\

     \makecell{EventScape\\~\cite{EventScape}} &
     %\makecell{Synthetic\\Dataset} &
     \includegraphics[width=0.95\linewidth]{dataset/Blender.png} &
     \includegraphics[width=0.95\linewidth]{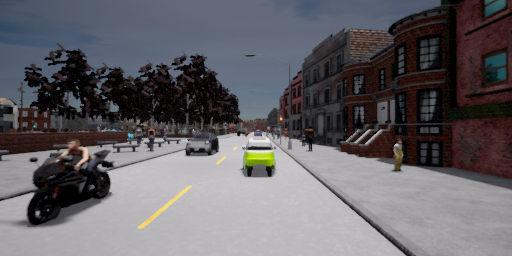} &
     \includegraphics[width=0.95\linewidth]{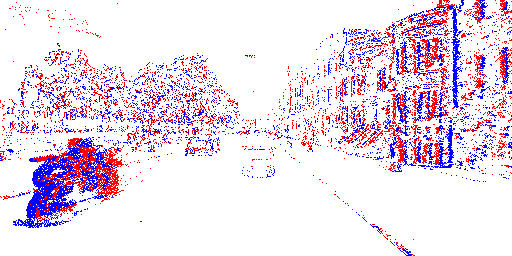} &
     \includegraphics[width=0.95\linewidth]{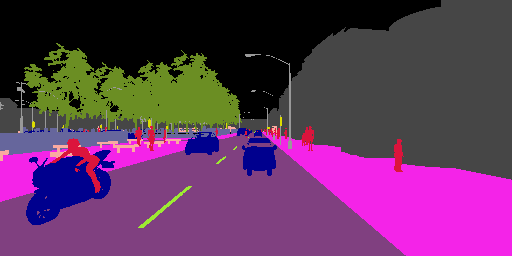} \\

     \makecell{SEVD\\~\cite{SEVD}} &
     %\makecell{Synthetic\\Dataset} &
     \includegraphics[width=0.95\linewidth]{dataset/Blender.png} &
     \includegraphics[width=0.95\linewidth]{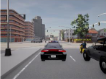} &
     \includegraphics[width=0.95\linewidth]{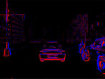} &
     \includegraphics[width=0.95\linewidth]{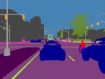} \\

     \bottomrule
   \end{tabular}
  }
  \end{adjustbox}
\end{table*}

\subsection{Photometric benchmarks}  
These datasets forego metric depth while delivering sharp RGB imagery with precise camera poses. EventNeRF \cite{[N3]} aligns asynchronous events with high-quality references for radiance-field research, while the classical DAVIS-240C sequences \cite{DAVIS240C} continue to serve as a compact baseline for structure-from-motion and visual-odometry modules.

\section{Metrics} \label{metrics}
Appropriate evaluation metrics are essential for fair comparison and objective analysis of event-based 3D reconstruction methods. Since different approaches vary in reconstruction targets and output representations, no single metric can fully characterize reconstruction quality. As a result, existing works adopt evaluation criteria that are tailored to specific method categories and output forms. In this section, we summarise the commonly used evaluation metrics and organize them according to different classes of event-based 3D reconstruction methods.

For traditional stereo and monocular geometry-based methods, multimodal methods with traditional outputs, the primary focus lies on the geometric accuracy of the reconstructed point clouds or meshes. Metrics such as Mean Error, Median Error, Relative Error, Root Mean Square Error (RMSE), and Mean Absolute Error (MAE) are employed to quantify the deviation in spatial occupancy and geometric consistency between the reconstruction results and the ground truth \cite{[S0],[S2],[S4],[XS2],[S6],[S7],[S8],[XS3],[XS4],[EMVSconf],[M3],[M4],[XDMT3],[XDMT1],[DMT2],[XDMT2],[DMT4],[DMT5],[DMT6]}. For monocular learning-based methods, Intersection over Union (IoU) and F-Score are adopted as the principal metrics to measure the consistency of object shape recovery \cite{[MD1],[Dense],[Towards]}. IoU serves as the most intuitive indicator for measuring the degree of overlap between two volumes (the reconstructed volume and the ground truth volume), assessing the consistency of spatial occupancy between the object shape recovered by the algorithm and the real object. The F-score is the harmonic mean of precision and recall; it requires not only accuracy in the reconstructed points but also completeness in the reconstruction, thereby providing a better comprehensive evaluation of the recovery effects regarding object boundaries and details.

Regarding neural rendering and novel view synthesis methods such as NeRF and 3DGS \cite{[N3],[N2],[XN2],[G10],[G5],[G8],[N3],[N14],[XG4],[XG6]}, the emphasis is placed on assessing the photometric fidelity and perceptual quality of the rendered images. Peak Signal-to-Noise Ratio (PSNR), Structural Similarity (SSIM), and Learned Perceptual Image Patch Similarity (LPIPS) are utilized as the core evaluation standards. PSNR reflects the purity of the image at the signal level, where a higher PSNR value indicates smaller pixel value errors and less distortion. SSIM attempts to simulate the human eye's perception capability regarding image structure. Unlike PSNR, which focuses solely on pixel value differences, SSIM compares two images across three dimensions: luminance, contrast, and structure. It posits that the human eye is more adept at capturing structural information within an image, such as object contours and texture orientation. LPIPS leverages pre-trained deep neural networks (simulating the human visual cortex) to extract image features and calculates the distance between these features; a lower LPIPS value signifies that the rendered image appears more natural and realistic to the human eye.

\section{Research Gaps \& Future Direction}  \label{7}

Despite recent advances, event-driven 3D reconstruction still faces key challenges across simulation, evaluation, modelling, and deployment. 

\subsection{Standardised datasets and benchmarks}
Despite the rapid methodological progress in event-based 3D reconstruction, the field continues to suffer from a scarcity of large-scale, real-world, and openly accessible datasets explicitly designed for reconstruction tasks \cite{[R2], [R3]}.

Many existing works rely on private datasets or synthetic event generation pipelines \cite{[XS2],[M2],[XDMT3],[DMT4],[N12],[G1]}, which significantly hinders fair comparison, reproducibility, and long-term benchmarking across methods. Publicly available datasets are often limited in scene diversity, confined to controlled laboratory environments, or released without dense geometric supervision, referring to Section \ref{6}.

To address this bottleneck, several community-level initiatives and collaborative efforts are essential.
First, standardised data acquisition pipelines that integrate event cameras with mature sensing modalities—such as LiDAR, RGB-D sensors, and motion-capture systems (e.g., Vicon) - should be promoted to reduce the technical and logistical barriers to dataset construction while ensuring high-quality geometric ground truth.
Second, multi-institution collaborations, similar to those established in autonomous driving and robotics research, would enable the collection of geographically and scenically diverse datasets under unified calibration, annotation, and synchronization protocols.
Third, the establishment of shared benchmarks, challenges, and public leaderboards, analogous to KITTI \cite{Geiger2012KITTI} or TUM \cite{Sturm2012TUM} in conventional vision, could provide strong incentives for open data release, sustained dataset maintenance, and transparent method comparison.
Finally, physically grounded simulators and differentiable event-generation models offer a complementary pathway to augment real-world data \cite{Hu2021V2E, [esim]}, provided that domain gaps between synthetic and real events are systematically quantified and mitigated through hybrid real–synthetic evaluation protocols.

These initiatives are critical for transforming event-based 3D reconstruction from a collection of isolated, dataset-specific studies into a unified, reproducible, and scalable research ecosystem.

\subsection{Synthetic event datasets not reliant on frame interpolation}
Currently, the main modelling platforms do not support event-based 3D modelling. While simulation tools such as ESIM \cite{[esim]} and Video-to-Event \cite{[v2e]} exist, the event streams they generate rely on brightness changes between consecutive image frames, which introduces noticeable frame-based artefacts and fails to capture the sparse and asynchronous nature of real event data. 

The latest simulator for event cameras is showing the possibilities of creating such datasets \cite{eventSimulator, eventSimulator2}. In the future, simulating event cameras within 3D modelling environments, which enables the collection of highly realistic event streams with perfect ground truth, holds great potential for deep learning in 3D reconstruction tasks.

\subsection{Event representation}
Event representation remains one of the most critical and promising research directions in event-based 3D reconstruction. Existing studies have shown that the choice of event representation has a substantial impact on feature extraction and learning behavior. However, its influence on reconstruction accuracy, robustness, and computational efficiency has not yet been systematically explored. Although a variety of event representations have been proposed in recent years \cite{[Eventrep1],[Eventrep2],[Towards]}, there is still a lack of comprehensive comparison under a unified reconstruction framework and evaluation protocol, which limits a clear understanding of their relative strengths and weaknesses.

Future research should first focus on conducting rigorous evaluations of representative event encodings within a unified 3D reconstruction pipeline. Such studies should analyze their performance in terms of geometric accuracy, temporal consistency, robustness, and computational cost, particularly across different scenarios such as static and dynamic scenes, high-speed motion, low-texture environments, and challenging illumination conditions. This would provide task-driven guidance for selecting appropriate event representations in practical 3D reconstruction systems.
Beyond fixed representations, a particularly promising direction lies in adaptive and hybrid event representations. Instead of relying on predefined temporal windows or spatial aggregation strategies, these approaches can dynamically adjust their spatiotemporal encoding according to scene dynamics, event density, motion intensity, or task requirements. For example, high temporal resolution representations may be favored in rapidly moving regions, while stronger spatial aggregation can be applied to more stable structures, potentially achieving a better balance between reconstruction accuracy and computational efficiency.

In addition, developing geometry-aware event representations constitutes an important research avenue. Compared to generic event encodings, such representations can explicitly emphasize geometrically informative cues such as edges, surface discontinuities, and structural consistency, thereby providing more discriminative inputs for downstream tasks including depth estimation, volumetric reconstruction, and neural rendering.

\subsection{Real-time reconstruction of dynamic scenes}
Most existing event-driven 3D reconstruction methods primarily focus on static scenes, while extending them to dynamic environments introduces substantial challenges in both modelling and computation \cite{[R1], [EMVS]}. In particular, achieving real-time performance alongside high-fidelity reconstruction remains challenging for neural-rendering-based approaches, especially in the presence of non-rigid motion and temporal variation \cite{Pumarola2021DNeRF}. Rather than treating dynamic reconstruction as a direct extension of static pipelines, recent studies suggest that explicitly modelling temporal dynamics constitutes a more promising research direction.

Future research should therefore focus on representations and models that explicitly encode motion and temporal evolution, such as deformable Neural Radiance Fields or dynamic 3D Gaussian Splatting augmented with temporal embeddings or motion-aware primitives \cite{Pumarola2021DNeRF}. 

From a real-time perspective, another promising avenue lies in prioritizing spatial occupancy and coarse geometry over photorealistic appearance. For many downstream applications such as robotics, augmented reality, and autonomous driving, accurate reconstruction of scene structure and free space is often more critical than detailed texture or colour reproduction \cite{[EMVS], [M1]}. This observation motivates lightweight representations that focus on dynamic occupancy, depth, or surface geometry, while simplifying or deferring appearance modelling.

The most promising directions for real-time dynamic scene reconstruction are likely to emerge from the combination of motion-aware representations, event-driven temporal modelling, and task-oriented simplifications that balance reconstruction fidelity with computational efficiency.

\subsection{Experiments under extreme scenarios}
It is well known that event cameras perform exceptionally under extreme conditions such as high-speed motion, low illumination, and high dynamic range. However, in the field of 3D reconstruction, comprehensive benchmarking against traditional cameras under such conditions remains limited \cite{[Towards]}. Future work should design experiments to systematically evaluate and enhance the robustness of event cameras in these challenging scenarios. This may involve developing novel algorithms or sensor fusion strategies to achieve reliable reconstruction in environments that are difficult for conventional visual sensors.

\subsection{Reconstruction of object with challenging materials}
Reconstructing 3D scenes with low-texture or non-Lambertian surfaces \cite{zhang2004shape, li2020inverse} (e.g., glass or mirrors) remains highly challenging and is rarely studied. While the event camera’s sensitivity to photometric changes could potentially complement traditional cameras in such tasks, it may also be adversely affected. To date, only a few studies have explored using event cameras for photometric stereo of these types of objects \cite{[EventPS]}. In addition, dynamic and non-rigid objects introduce further complexity \cite{Non-Rigid}, as their continuously changing geometry violates the assumptions of many static reconstruction methods. Recent works in event-based dynamic NeRF \cite{[N4],[N14]} show potential in this direction, but general solutions remain underexplored. Future research should investigate the effectiveness of event cameras in this context and, where applicable, design specialised algorithms tailored to reconstruct these challenging materials.

\subsection{Hardware and synchronisation constraints}
The performance of event-driven pipelines is often bounded by hardware limitations, including the event camera’s resolution, timestamp precision, and bandwidth. In multimodal systems, accurate synchronisation between events, frames, IMU, and structured light sources is difficult to achieve and critical for fusion-based reconstruction \cite{10901942}. Even small pose errors can propagate and degrade 3D structure estimation \cite{[S8]}.

\subsection{Efficiency and scalability bottlenecks}
NeRF and 3DGS methods offer high-fidelity reconstructions but are computationally expensive and memory-intensive. Real-time or large-scale scene reconstruction remains impractical, particularly in mobile or embedded scenarios. Efforts such as Event3DGS \cite{[G4]} and EVI-SAM \cite{[M3]} have started to address this, but further work is required on efficient architectures, pruning strategies, and event-specific network designs.

\subsection{Underexplored modalities}
While structured light and RGB-D sensors have been successfully combined with event cameras \cite{li2022high}, other modalities such as LiDAR, polarisation cameras, and event-based time-of-flight sensors remain comparatively underexplored. Their integration could introduce complementary geometric and physical priors, particularly under fast motion or challenging illumination.

In multimodal 3D reconstruction, effective fusion strategies focus on complementary sensing rather than simple data aggregation. Event cameras provide high-temporal-resolution motion cues but lack spatial density and absolute scale, whereas LiDAR and RGB-D sensors offer metric geometric constraints but are sensitive to motion artifacts, and IMU measurements ensure short-term motion consistency at the cost of long-term drift. 

Promising approaches therefore perform fusion at both the representation and optimization levels. At the front end, accurate temporal synchronization enables events to capture rapid motion while depth or LiDAR measurements act as stable geometric anchors and IMU signals constrain pose continuity. At the back end, joint optimization or neural representation frameworks, such as multimodal NeRF or 3D Gaussian Splatting, map heterogeneous observations into a unified 3D representation with modality-specific losses and adaptive weighting to suppress sensor-dependent noise.

Importantly, selective modality activation based on scene dynamics represents a particularly effective strategy. For example, event data can be emphasized in high-speed or high-dynamic-range scenarios, while depth or LiDAR cues dominate in structurally stable regions, enabling improved reconstruction quality and real-time performance \cite{[DMT5]}.

\subsection{Exploring broader downstream applications}
Due to the limitations of existing datasets, current research on event-based 3D reconstruction remains constrained to scenarios supported by available benchmarks. However, there are many foreseeable downstream applications yet to be fully explored: 
\begin{itemize}
    \item UAVs: Mounting event cameras on unmanned aerial vehicles (UAVs) could enable the reconstruction of large-scale objects or architectural structures under challenging conditions such as high-speed motion or dynamic lighting.
    \item Robotics: Event-based 3D reconstruction holds great promise in robotics for fast, low-latency perception and navigation in cluttered or fast-changing environments.
    \item Autonomous Driving: Event cameras enable stable 3D perception in scenarios where conventional cameras struggle, such as nighttime, tunnels, or strong backlighting. In autonomous driving, they offer reliable geometric sensing under high dynamic range conditions or rapid motion, serving as a robust complement to LiDAR and standard cameras.
    \item VR: In augmented and virtual reality, event-driven depth sensing may enable energy-efficient and low-latency 3D interaction.
    \item Cultural Heritage Scanning: Event cameras may enable non-invasive 3D reconstruction of fragile artefacts and heritage sites under low-light or vibration-sensitive conditions, offering a safe alternative for digitising and monitoring valuable cultural assets.
    \item Other industrial applications, such as quality inspection of reflective or high-speed moving parts, as well as medical imaging under low-light or non-invasive conditions, represent other promising directions.
\end{itemize}  

\section{Conclusion}

This survey presents a comprehensive and systematic review of 3D reconstruction techniques based on event cameras, which are emerging as a powerful alternative to conventional vision sensors in challenging environments. We categorised the literature by input modality, including stereo, monocular, and multimodal systems, and further by reconstruction strategy, ranging from geometry-based and deep learning-based pipelines to recent advances in neural rendering using Neural Radiance Fields and 3D Gaussian Splatting. Methods with a similar research focus were organised chronologically into the most subdivided groups. Through detailed comparisons and timeline visualisations, we revealed the evolution and diversification of event-driven 3D reconstruction. We also compiled a list of publicly available datasets to support reproducibility and benchmarking. Despite the progress, we identified critical research gaps in dataset standardisation, event representation design, dynamic scene modelling, real-time deployment, etc. As event cameras continue to mature, we anticipate further breakthroughs in both theoretical modelling and practical applications, particularly under extreme motion and illumination conditions. We hope this survey provides a useful foundation for new researchers and a roadmap for advancing the field of event-based 3D reconstruction.

\subsection*{Author contributions}

Chuanzhi Xu proposed the review topic, collected most of the referenced papers, and conducted and structured the majority of the work. He also created the majority of the tables and figures. Haoxian Zhou contributed significantly to the writing of several sections and produced multiple figures. Langyi Chen wrote Section 6 on datasets. Haodong Chen, Zeke Zexi Hu, and Zhicheng Lu carefully reviewed and revised the paper. Qiang Qu coordinated the project and managed subsequent communications. Ying Zhou, Vera Chung, and Weidong Cai supervised the project, provided necessary materials, and offered general guidance and feedback on the manuscript.

\subsection*{Declaration of competing interest}

The authors have no competing interests to declare that are relevant to the content of this article.

% for bibtex
\bibliographystyle{CVMbib}
\bibliography{refs}

\end{document}